\documentclass[journal,twoside]{IEEEtran}
\usepackage{amsmath,amsfonts}
\usepackage{algorithmic}
\usepackage{algorithm}
\usepackage{array}
\usepackage[caption=false,font=normalsize,labelfont=sf,textfont=sf]{subfig}
\usepackage{textcomp}
\usepackage{stfloats}
\usepackage{url}
\usepackage{verbatim}
\usepackage{graphicx}
\usepackage{cite}
\usepackage[table]{xcolor}  %表格颜色
\usepackage{booktabs}
\usepackage{adjustbox}
\usepackage{multirow}
\usepackage{tabularx} %表格
\newcolumntype{Y}{>{\centering\arraybackslash}X}
\usepackage{hyperref}
\hypersetup{colorlinks=true, linkcolor=blue, citecolor=blue, urlcolor=black}
\usepackage{amssymb}
\usepackage{ragged2e}
\usepackage{makecell} %单元格跨行
\usepackage{threeparttable} %表格注释
\definecolor{lightblue}{RGB}{60, 39, 139}

\begin{document}

\title{KILVO: Kinematic-Inertial-LiDAR-Visual Odometry with Robust Multimodal Adaptation for Humanoid Robots}

\author{Jixin Gao, \IEEEmembership{Student Member, IEEE}, Fucheng Liu, Teng Zhang, and Fusheng Zha

\thanks{
  This work was supported in part by the Suzhou Science and Technology Research under Grant SYG2025113,
          in part by the Key Technology Research and Development of Humanoid Robots under Grant SYG2024142,
          and in part by the Self-Planned Task of State Key Laboratory of Robotics and Systems (HIT) under Grant 2023FRFK01001.
  \textit{(Corresponding author: Fusheng Zha.)}
}% <-this % stops a space
\thanks{The authors are with the State Key Laboratory of Robotics and Systems, Harbin Institute of Technology, Harbin 150006, China (e-mail: 24b936018@stu.hit.edu.cn; 23b908027@stu.hit.edu.cn; 24b908047@stu.hit.edu.cn; zhafusheng@hit.edu.cn).}
\thanks{\protect\url{https://github.com/JixinGao/KILVO}}
\thanks{Digital Object Identifier (DOI  ): 10.1109/TMECH.2026.3721778.}
}

\markboth{IEEE/ASME TRANSACTIONS ON MECHATRONICS}%
{Gao \MakeLowercase{et al.}: KILVO: Kinematic-Inertial-LiDAR-Visual Odometry with Robust Multimodal Adaptation for Humanoid Robots}

\IEEEpubid{}

\maketitle

\begin{abstract}
This article presents a kinematic-inertial-LiDAR-visual odometry for humanoid robots, called KILVO.
Tailored to the platform features, requirements, and real-world complexity, 
it fully utilizes the sensors commonly equipped on humanoid robots, 
including joint encoders, IMU, LiDAR, and camera,
within an asynchronous-sequential hybrid error-state iterated Kalman filter (ESIKF).
Specifically, inertial data are used for prediction, leg kinematics are processed asynchronously at a high rate and provide proprioceptive constraints, while exteroception is updated sequentially, first by registering LiDAR points for geometric priors and then by updating the visual component via photometric errors.
Moreover, the framework is elaborately designed with multimodal adaptation for resilience to sensor failures.
A compact contact estimation module is also developed, sharing information with state estimation without additional sensors.
Extensive experiments on public datasets and in the real world across multiple humanoid robots, gait patterns, and scenarios
demonstrate that KILVO achieves highly competitive accuracy, efficiency, and output rates, with strong robustness against sensor degradation and failures, making it more suitable for humanoid robots than state-of-the-art fusion methods.
Our code and datasets are released on GitHub.
\end{abstract}


\section{Introduction}
\IEEEPARstart{R}{ecently}, humanoid robots are becoming increasingly capable, exhibiting immense potential for real-world applications and complex task execution \cite{JAS_2024}.
Odometry is a key technique for these robots to estimate poses and reconstruct maps in unknown environments.
It has become indispensable in providing the crucial information for high-level tasks such as exploration and planning, as well as low-level functions like locomotion controllers.

Most legged robots are equipped with IMUs, joint encoders, and contact sensors to enable state estimation for control and navigation, also known as proprioceptive odometry.
Typically implemented with nonlinear filters, such methods provide high-rate state feedback with low computational load \cite{2014_po, IJRR_InEKF}.
However, they tend to accumulate severe drift under non-ideal conditions (e.g., challenging terrains, dynamic motions, long-term tasks).
This issue is particularly pronounced on humanoid robots, {whose} vertical chain structure, {low-redundancy contacts}, and growing locomotion capability {further amplify irregular measurement noise and constraint uncertainty.}
\begin{figure}[!t]
\centering
\includegraphics[width=\linewidth]{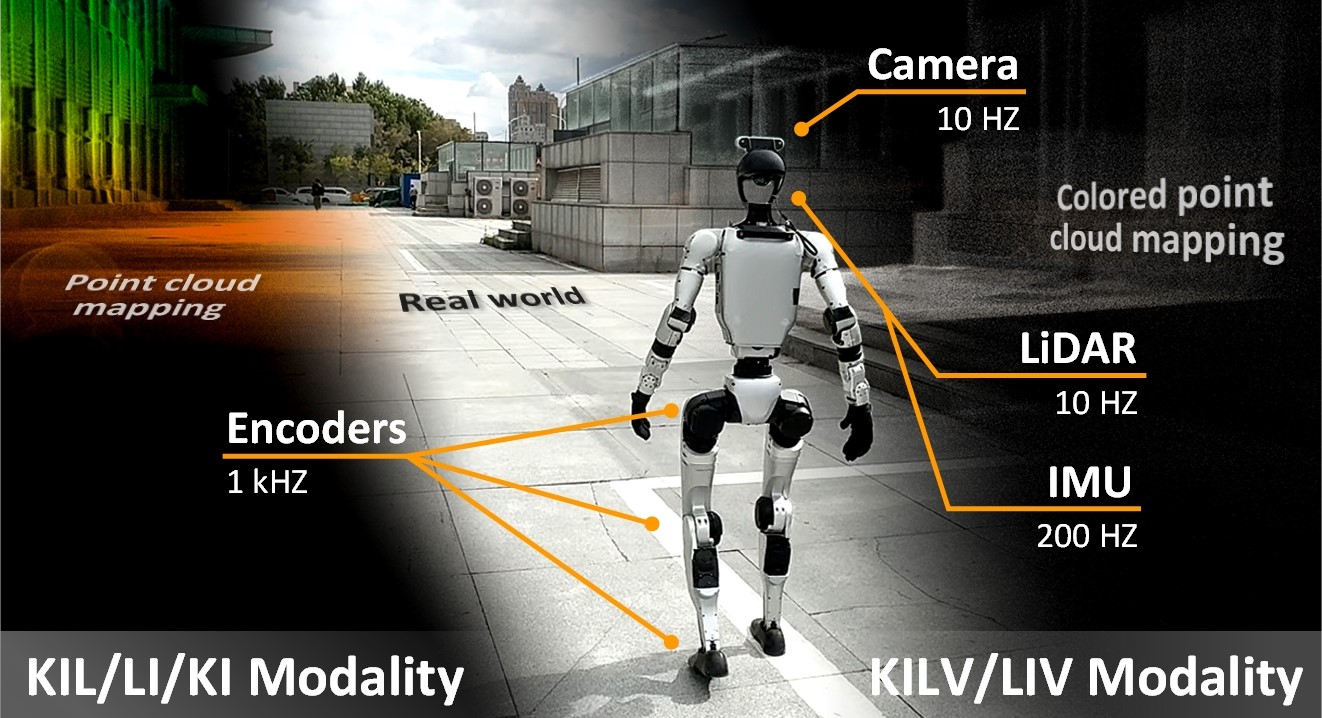}
\caption{KILVO leverages standard onboard sensors on humanoid robots to deliver accurate, efficient localization at 1 kHz and mapping with colored points, while enabling seamless modality adaptation to tolerate sensor failures.}
\label{fig_1}
\end{figure}

With the advancement in simultaneous localization and mapping (SLAM), fusion of commonly used exteroceptive sensors is becoming a prominent solution.
LiDAR and camera are fused with proprioception in recent works for accurate state estimation of humanoid robots \cite{RoboSLAM,GeoFlow2025, 2014LiDAR}.
The high-rate and rigorous real-time {requirements} are also focused in \cite{LIKO,HR2-KILO}.
Although these fusion approaches have advanced the field, it remains difficult to provide reliable support for humanoid robots' anytime-anywhere autonomy.
This is because the instability of humanoid platforms and {real-world} complexity make sensor degradation and failure possible at any time, 
% which is subject to the dynamic motions, unexpected falls/collisions, and harsh environments.
{such as body vibrations aggravating data association, measurement degradation introducing unreliable constraints, and even data interruptions caused by} collisions/falls.
Non-resilient systems {in tightly coupled pipelines typically assume the continuity of sensing modalities and lack the } multimodal adaptation {to respond to} the above risks, leading to potential task failures or even robot damage.
Additionally, reliable contact information is essential for kinematic updates \cite{contact_learn_2021}.
Yet most systems obtain it in an additional manner, relying on dedicated sensors, ground reaction forces (GRF) thresholding, or learning-based approaches, each with limitations that further pose a challenge for widespread deployment.

To address these challenges, this article presents KILVO, a multisensor fusion odometry for humanoid robots that integrates joint encoder, IMU, LiDAR, and camera.
{Considering the heterogeneous measurements,} we perform state updates through an asynchronous-sequential hybrid error-state iterated Kalman filter (ESIKF), {which tightly couples multisource data while maintaining efficient, high-rate outputs}.
{This also facilitates} the framework{'s robust adaptation} across modalities, {thereby tolerating} sensor failures.
For contact estimation, we {reuse} kinematic, inertial, and map cues, allowing compact system integration and {avoiding the hardware dependency}.
The main contributions of this article are as follows.
\begin{enumerate}
  \item{
    A novel multisensor framework for state estimation on humanoid robots.
    The system is {elaborately designed to} tightly couple leg kinematic, inertial, LiDAR, and visual data with asynchronous-sequential hybrid update in ESIKF, 
    while supporting seamless multimodal adaptation to handle sensor failures, as illustrated in Fig \ref{fig_1}. 
  }
  \item{
    Stable and efficient contact estimation without additional sensors.
    The module detects bipedal contact states by calculating impact deviation and foot clearance based on existing data.
    Impact deviation is also reused for modality {adaptation} and leg kinematic update.
  }
  \item{
    A comprehensive SLAM dataset for humanoid robots.
    We collected leg kinematic, inertial, LiDAR, and image data, along with relative ground truth for the dataset.
    A total of 15 sequences cover different gaits, sensor-degradation scenarios, and diverse terrains.
  }
  \item{
    Extensive evaluation including public datasets and real-world scenarios with different configurations. 
    The results demonstrate our system achieves competitive accuracy with 1 kHz output, also exhibits strong robustness to resist sensor degradation and failures.
    The code and datasets of our work are released to the community.
  }
  % \item{
  %   The code and datasets of KILVO are released to the community to promote the reproducibility and further development of our work.
  % }
\end{enumerate}
\section{Related Works}
\subsection{Odometry on Humanoid Robots}
The earlier work \cite{2014_po} extended the StarETH state estimator \cite{2012_starlETH} to humanoid robots by leveraging proprioceptive sensing within an error-state extended Kalman filter (ESKF).
Similarly, the works \cite{2023IROS_IMUs,2022RAL_IMUs}, also based on ESKF, used multiple IMUs to jointly estimate the body state and structural deformation.
For better convergence properties, the invariant observer was introduced in \cite{IJRR_InEKF} and validated on Cassie-series bipedal robots, inspiring the subsequent proprioceptive odometry \cite{2021DILIGENT_KIO_RAL,2022Tmech_GY,2025Tmech_GY}.
Such proprioceptive-only methods serve as a low-cost and efficient solution, but are typically limited by the unobservability of yaw and position, as discussed in \cite{2012_starlETH,IJRR_InEKF}.

The incorporation of exteroception has gained increasing attention, especially camera and LiDAR.
An early work \cite{2006IROS_MonoSLAM_CPG} introduced a pattern generator in SLAM for humanoid robots.
As these robots become less bound by fixed patterns, the community has focused on the compound challenges of illumination variations and motion blur, including feature-based methods \cite{2025CAC_cams,GeoFlow2025} and direct methods \cite{2016Direct,2017Direct,RoboSLAM}.
However, vision-based systems are still limited by computational overhead, illumination sensitivity, and textureless scenes.
LiDAR was probabilistically fused with inertial and kinematic data in \cite{2014LiDAR} to infer Atlas's position relative to a prior map.
Recently, LIKO \cite{LIKO} and our previous work HR$^2$-KILO \cite{HR2-KILO} integrated leg kinematics with LIO for high-rate and robust state estimation.
Despite these advances, few works tightly couple various sensors (encoder, IMU, LiDAR, and camera) commonly equipped on humanoid robots to {provide complementary constraints for accurate and high-rate state estimation.}

\subsection{SLAM with Multiple Modalities}
The system for humanoid robots is also expected to flexibly adapt to sensor failures, so we briefly review multimodal SLAM in this section.
{R$^3$LIVE \cite{R3LIVE} employs collaborative LIO/VIO subsystems based on ESIKF to improve robustness in degraded environments.}
In \cite{LIV-SAM, VILENS, LVI-Q}, {LiDAR, camera, IMU, and even joint encoder} are tightly integrated {based on hybrid/factor-graph optimization (FGO)}. 
{The above system resilience is mainly focused on measurement degradation and constraint availability, rather than data interruption and the resulting estimator reconfiguration.}
The works \cite{MIMOSA,LOCUS2} achieved robustness to temporary data loss via a health monitor, 
and the recent learning-based odometry \cite{SCINECE_SuperOdo} introduced hierarchical adaptation {within the FGO framework}.
{These works show a progression from measurement robustness to system-level resilience.}
However, the general methods are not designed for humanoid robots, and we prioritize an efficient filter-based pipeline over {FGO}, although it can process different modalities more flexibly.

\subsection{Contact Estimation}
Contact events are commonly detected by empirically thresholding GRFs based on dynamic models, as used in \cite{contact_TO_2017,contact_ProEsti_2020}.
This is effective in laboratories but is limited in robustness across various terrains and gait patterns.
In \cite{2016_contact}, leg kinematics were additionally fused for probabilistic contact estimation.
Similarly, probabilistic contact also includes \cite{2023_contact}, which used IMUs mounted on the robot's end effectors.
Lin et al. \cite{contact_learn_2021} developed a deep-learning-based method to classify the individual contacts for each foot.
By utilizing IMUs and force/torque (F/T) measurements from legs, the model in \cite{2022_contact_learning} {was validated} across different robotic platforms.
{But compared with analytical methods, they still depend on the training-data coverage and cost}.
In HR$^2$-KILO \cite{HR2-KILO}, contact estimation was integrated with the state estimator by sharing available information, making the system compact and easy to deploy. 
While it suffers from {several} limitations, {such as} coupled estimation of biped contacts, detection delays, and {dependence on LIO}, {which} are addressed in this article.
\section{System Overview}
\begin{figure*}[!t]
\centering
\includegraphics[width=\linewidth]{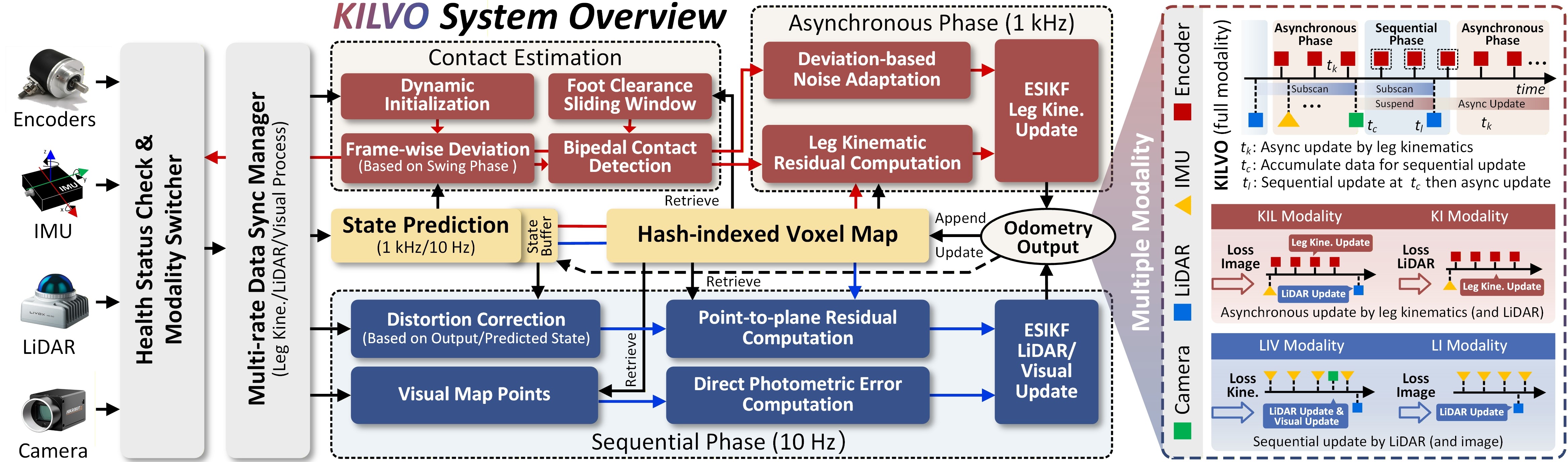}
\caption{System overview of KILVO and the processing for each modality.}
\label{fig_2}
\end{figure*}

The system overview is shown in Fig. \ref{fig_2}.
The measurements from joint encoders, IMU, LiDAR, and camera are fed into the system at their respective frequencies and first checked for data {health status}.
When all the sensor data are valid, we tightly couple them via an ESIKF with asynchronous-sequential hybrid updates,
in which IMU data is used for state prediction, leg kinematics is for asynchronous state updates, while LiDAR and visual updates are performed sequentially.
Specifically, in the asynchronous phase (1 kHz), we construct multi-constraint leg kinematic residuals and {update the state with deviation-based} noise adaptation.
The updated states are directly output and cached in a state buffer.
In the sequential phase (10 Hz), we undistort the LiDAR points by the state buffer, followed by point-to-plane residual calculation and LiDAR update.
Then, extract visual map points and calculate frame-to-map photometric errors for visual update.  
For these residual constructions, we deploy a unified hash-indexed voxel map that also serves contact estimation. 
The points after LiDAR update are appended to the map to update the geometric structure, while images following the visual update are used to update the reference patches for visual map points.

A contact estimation module driven by existing data is also developed (Section \ref{sec:5}). 
For dynamic initialization, inertial and kinematic data from system startup and the first swing phase are used to establish reference statistics.
We then calculate the normalized deviation frame-by-frame, while retrieving the ground patch from the map to build a clearance sliding window.
{The obtained contact states and deviations are further used for the kinematic update with its covariance reweighting.}

To fully realize the potential of multisensor fusion, the modality switcher and data synchronization manager are involved.
This enables KILVO to robustly adapt to available modalities during sensor failures, thereby maintaining state consistency. 
The processing of these modalities is shown on the right side of Fig. \ref{fig_2}, detailed in Section \ref{sec:4-e}. 

\section{ESIKF with Hybrid Process}

\subsection{Definitions}
  In this article, 
  all the sensors are treated as rigidly attached with precalibrated extrinsics, and the temporal alignment is not necessary.
  We assume the IMU frame coincides with the robot body frame, its first frame is considered the world frame. 
  Relevant notations can be found in Table \ref{tab:1notation}.
  \begin{table}[t!]
    \caption{Some Notations in This Article\label{tab:1notation}} %------------------------------------ TABLE BEG
    \centering
    {
      \newcolumntype{L}{>{\RaggedRight\hangafter=1\hangindent=0em}X}
      \begin{tabularx}{\linewidth}{p{2.3cm} L}
        \toprule
        Notations & Explanations\\
        \midrule[0.5pt]
        ${}^I(\cdot), {}^L(\cdot)$, and ${}^C(\cdot)$ & IMU frame, LiDAR frame, and camera frame.\\
        ${}^W(\cdot)$                                 & A vector in global world frame.   \\
        $\lfloor \cdot \rfloor _\wedge$               & Skew-symmetric cross product matrix. \\
        $\boxplus$ / $\boxminus$                      & Encapsulation operators on the state manifold.  \\
        $\operatorname{Exp(\cdot)}$ / $\operatorname{Log(\cdot)}$ & Bidirectional mapping between the rotation matrix and rotation vector.\\
        ${}^L\mathbf{T}_{I}$ and ${}^C\mathbf{T}_{L}$ & Extrinsic between different sensors. \\
        ${}^W\mathbf{T}_{I}$ and ${}^W\mathbf{T}_{L}$ & Transformation from IMU frame/LiDAR frame to the world frame. \\
        $\mathbf{x}, \hat{\mathbf{x}}$, and $\bar{\mathbf{x}}$   & Ground-true, predicted, and updated states.  \\
        $\tilde{\mathbf{x}}$ and $\hat{\mathbf{x}}^{\kappa}$       & Error state and updated state at the $\kappa$th iteration.\\
        \bottomrule
      \end{tabularx}
    }
  \end{table} %------------------------------------------------------------------------- TABLE END
  Then, the state vector $\mathbf{x}$,  input $\mathbf{u}$, and  process noise $\mathbf{w}$ are defined as:
  \begin{equation}
    \begin{aligned}
      \mathcal{M}&\triangleq \mathrm{SO}(3)  \times \mathbb{R}^{19}, dim(\mathcal{M}) = 22 \\
      \mathbf{x} &\triangleq 
      \begin{bmatrix}
        {}^W \mathbf{R}^T_I \ {}^W \mathbf{p}^T_I \ {}^W \mathbf{v}^T_I \ \mathbf{b}^T_{g} \ \mathbf{b}^T_{a} \ {}^W\mathbf{g}^T \ {}^W\mathbf{p}^T_c \ \tau
      \end{bmatrix} ^T \in \mathcal{M}\\
      \mathbf{u} &\triangleq \begin{bmatrix} \boldsymbol{\omega}^T_m \ \mathbf{a}^T_m \end{bmatrix} ^T,
      \mathbf{w} \triangleq
      \begin{bmatrix}
        \mathbf{n}^T_{g} \ \mathbf{n}^T_a \ \mathbf{n}^T_{\mathbf{b}_g} \ \mathbf{n}^T_{\mathbf{b}_a} \ \mathbf{n}^T_{\mathbf{p}_c} \ n_{\tau}
      \end{bmatrix} ^T
    \end{aligned}
  \end{equation}
  where ${}^W \mathbf{R}_I$, ${}^W \mathbf{p}_I$, and ${}^W \mathbf{v}_I$ are the body rotation, position, and velocity,
  $\mathbf{b}_{g}$ and $\mathbf{b}_{a}$ are the IMU bias,
  ${}^W\mathbf{p}_c$ is the contact foot position, ${}^W\mathbf{g}$ is the gravity vector,
  $\tau$ is the  inverse camera exposure time, % relative to the first frame, 
  $\boldsymbol{\omega}_m$ and $\mathbf{a}_m$ are the {angular velocity} and {acceleration} measured by the IMU,
  $\mathbf{w}$ is the process noise including measurement noises $\mathbf{n}_{g}$ and $\mathbf{n}_a$, bias noises $\mathbf{n}_{\mathbf{b}_g}$ and $\mathbf{n}_{\mathbf{b}_a}$, contact noise $\mathbf{n}_{\mathbf{p}_c}$, and exposure time noise $n_{\tau}$.

\subsection{State Prediction}   %--------------------------------------------------PREDICT
  In the full modality (all the measurements are valid), the system keeps at least one latest IMU measurement for state prediction aligned with the high-frequency update.
  Assume the prediction lasts from $t_{i}$ to $t_{i+1}$, the discretized process model is as follows:
  \begin{equation}
		\begin{gathered}
      \mathbf{x}_{i+1} =\mathbf{x}_i\boxplus(\Delta t\mathbf{f}(\mathbf{x}_i, \mathbf{u}_i, \mathbf{w}_i)), \\
      \mathbf{f} = 
        \begin{bmatrix}
          \boldsymbol{\omega}_{m_i} - \mathbf{b}_{g_i} - \mathbf{n}_{g_i} \\
          {}^W\mathbf{v}_{I_i} + \frac{1}{2}({}^W\mathbf{R}_{I_i} (\mathbf{a}_{m_i} - \mathbf{b}_{a_i} - \mathbf{n}_{a_i}) + {}^W\mathbf{g}_i)\Delta t \\
          {}^W\mathbf{R}_{I_i} (\mathbf{a}_{m_i} - \mathbf{b}_{a_i}- \mathbf{n}_{a_i})  + {}^W\mathbf{g}_i \\
          \mathbf{n}_{{\mathbf{b}_{gi}}} \\
          \mathbf{n}_{{\mathbf{b}}_{ai}} \\
          \mathbf{0}_{3 \times 1} \\
          {}^W\mathbf{R}_{I_i}\cdot\mathbf{h_R}(\boldsymbol{\alpha}_i)\cdot\mathbf{n}_{\mathbf{p}_{ci}} \\
          n_{\tau_i}
        \end{bmatrix} 
		\end{gathered}
    \label{eq:2}
  \end{equation}
  where $\mathbf{a}_{m_i}$ and $\boldsymbol{\omega}_{m_i}$ are from the latest available IMU data rather than strictly sampled at $t_{i}$,
  $\mathbf{h_R}(\boldsymbol{\alpha}_i)$ is the orientation from the foot contact frame to the IMU frame by forward kinematics,
  the notations $\boxplus / \boxminus$ in \cite{Boxplus} are introduced to compactly represent the state on manifold $\mathcal{M}$, and have:
  $$
  	\begin{aligned}
      \begin{bmatrix}\mathbf{R} \\ \mathbf{a}\end{bmatrix} \!\boxplus\! \begin{bmatrix}\mathbf{r} \\ \mathbf{b}\end{bmatrix} \!=\! \begin{bmatrix}\mathbf{R}\operatorname{Exp}(\mathbf{r}) \\ \mathbf{a}{+}\mathbf{b}\end{bmatrix},
      \begin{bmatrix}\mathbf{R}_1 \\ \mathbf{a}\end{bmatrix} \!\boxminus\! \begin{bmatrix}\mathbf{R}_2 \\ \mathbf{b}\end{bmatrix} \!=\! \begin{bmatrix}\operatorname{Log}(\mathbf{R}_2^T \mathbf{R}_1) \\ \mathbf{a}{-}\mathbf{b}\end{bmatrix}.
		\end{aligned}
  $$

  Following the process model in \eqref{eq:2}, the predicted state $\hat{\mathbf{x}}_{i+1}$ can be obtained by setting noise $\mathbf{w}_i$ to zero, 
  the predicted covariance $\hat{\mathbf{P}}_{i+1}$ is propagated as below:
  \begin{equation}
      \hat{\mathbf{P}}_{i+1} = \mathbf{F}_{\tilde{\mathbf{x}}_i} \hat{\mathbf{P}}_{i} \mathbf{F}^{T}_{\tilde{\mathbf{x}}_i} + \mathbf{F}_{\mathbf{w}_i} \mathbf{Q} \mathbf{F}^{T}_{\mathbf{w}_i}
  \label{eq:3}
  \end{equation}
  where $\mathbf{F}_{\tilde{\mathbf{x}}_i}$ and $\mathbf{F}_{\mathbf{w}_i}$ are the Jacobian matrices of $\tilde{\mathbf{x}}_{i+1}$ with respect to $\tilde{\mathbf{x}}_{i}$ and $\mathbf{w}_{i}$, respectively, 
  $\tilde{\mathbf{x}}_{i} \triangleq \mathbf{x}_i \boxminus \hat{\mathbf{x}}_i $, and $\mathbf{Q}_i$ is the covariance of $\mathbf{w}_i$.

\subsection{Asynchronous Phase} \label{sec:4-c}  %--------------------------------------------------ASYNC
\subsubsection{Kinematic Measurement}
  For {a foot} in contact with the ground, {as estimated} in Section \ref{sec:5}, we {construct kinematic measurements and} update the state in the asynchronous phase.
  Based on HR$^2$-KILO \cite{HR2-KILO}, three measurements of contact foot velocity ${\mathbf{h}_{cv}}$, position ${\mathbf{h}_{cp}}$, and height ${\mathbf{h}_{ch}}$ are considered:
  \begin{equation}
    \begin{aligned}
      \mathbf{0} &= \mathbf{h}_c(\mathbf{x}, \mathbf{n}_c)\\
                 &\triangleq    
      (
      \mathbf{h}_{cv}(\mathbf{x}, \mathbf{n}_{cv}),
      \mathbf{h}_{cp}(\mathbf{x}, \mathbf{n}_{cp}),
      \mathbf{h}_{ch}(\mathbf{x}, \mathbf{n}_{ch})
      )
    \end{aligned}
    \label{eq:leg_measurement_model}
  \end{equation}
  where $\mathbf{n}_{c} \sim \mathcal{N}(\mathbf{0}, \mathcal{R}_c^{nom})$ denotes the {kinematic} measurement noise {rather than the raw encoder noise}, {and it} is approximately modeled as three independent components with $\mathcal{R}_c^{nom} = \text{diag}(\mathcal{R}_{cv}, \mathcal{R}_{cp}, \mathcal{R}_{ch})$.

  Hence, the corresponding leg kinematic measurements are modeled as follows:
  \begin{equation}
    \begin{aligned}
      \mathbf{h}_{cv}(\mathbf{x}, \mathbf{n}_{cv}) &= -{}^W\mathbf{R}(\lfloor \boldsymbol{\omega} \rfloor _ \wedge {}^I\mathbf{p}_{fk} + {}^I \mathbf{v}_{fk}) - {}^W\mathbf{v} + \mathbf{n}_{cv}\\
      \mathbf{h}_{cp}(\mathbf{x}, \mathbf{n}_{cp}) &= {}^I\mathbf{p}_{fk} - {}^W \mathbf{R}^T ({}^W\hat{\mathbf{p}}_c - {}^W\mathbf{p}) + \mathbf{n}_{cp}\\
      \mathbf{h}_{ch}(\mathbf{x}, \mathbf{n}_{ch}) &= \mathbf{e}_z^T {}^W\mathbf{T}_{I} {}^I\mathbf{p}_{fk} - \mathbf{e}_z^T {}^W\mathbf{p}_{c_{proj}} + \mathbf{n}_{ch}
    \end{aligned}
  \end{equation}
  where ${}^I\mathbf{p}_{fk}$ and ${}^I \mathbf{v}_{fk}$ represent the measured foot position and velocity relative to IMU, derived from forward kinematics, $\mathbf{e}_z^T = [0, 0, 1]$, ${}^W\mathbf{p}_{c_{proj}}$ is the projection point of ${}^W\hat{\mathbf{p}}_c$ on the local ground patch, which is computed as follows:
  \begin{equation}
		\begin{aligned}
			{}^W\mathbf{p}_{c_{proj}} = {}^W\hat{\mathbf{p}}_{c} - {}^W\mathbf{u}^T({}^W\hat{\mathbf{p}}_{c} - {}^W\mathbf{q}){}^W\mathbf{u}
		\end{aligned}
  \end{equation}
  where the plane parameters $(\mathbf{u}, \mathbf{q})$ are from the voxel map or historical ${}^W\mathbf{p}_c$, depending on the modality of KILVO.

  \subsubsection{Iterative Update}
  Linearizing the kinematic measurement model (\ref{eq:leg_measurement_model}) through its first-order Taylor expansion at $\hat{\mathbf{x}}^{\kappa}$ leads to the residual $\mathbf{r}_c$:
  \begin{equation}
    \begin{aligned}
      \mathbf{r}_c =
      \mathbf{h}_{c}(\mathbf{x}, \mathbf{n}_{c}) - \mathbf{h}_{c}(\hat{\mathbf{x}}^{\kappa}, \mathbf{0})
      \simeq \mathbf{H}_{c}^{\kappa}\tilde{\mathbf{x}}^{\kappa} + \mathbf{D}_c^{\kappa} \mathbf{n}_c
    \end{aligned}
    \label{eq:leg_Resi}
  \end{equation}
  where $\mathbf{H}_c$ and $\mathbf{D}_c$ are the Jacobian matrices of residual $\mathbf{r}_c$ with respect to $\mathbf{\tilde{\mathbf{x}}}^{\kappa}$ and $\mathbf{n}_c$, respectively.

  To robustly handle ground impacts, we adaptively adjust $\mathcal{R}_{c}^{nom}$ in the state update.
  Specifically, a normalized deviation score $\mathbf{d}_r$ is calculated based on the current residual and the reference statistics $(\boldsymbol{\mu}_r^{sw}, \boldsymbol{\sigma}_r^{sw})$, {obtained from the contact estimation module}, detailed in Section \ref{sec:5-a}.
  The $\mathbf{d}_r$ is then used to re-weight $\mathcal{R}_{c}^{nom}$ after scaling with $\alpha_s$.
  This process can be summarized as follows:
  \begin{equation}
    \begin{aligned}
      \mathbf{d}_r = \left| \frac{\mathbf{r}_c - \boldsymbol{\mu}_r^{sw}}{\boldsymbol{\sigma}_r^{sw}} \right|, 
      \mathcal{R}_c = \mathrm{diag}\left( \frac{\alpha_s \mathbf{d}_r}{\mathrm{max}(\mathbf{d}_r)} \right)  \mathcal{R}_c^{nom}.
    \end{aligned}
    \label{eq:leg_noise_adp}
  \end{equation}

  Finally, by calculating the Kalman gain $\mathbf{K}_c$ in ESIKF framework, the posterior state can be obtained:
  \begin{equation}
    \begin{aligned}
      \mathbf{K}_c &= ({\mathbf{H}_c^{\kappa}}^T \mathcal{R}_c^{-1} \mathbf{H}_c^{\kappa} + \hat{\mathbf{P}}^{-1})^{-1} {\mathbf{H}_c^{\kappa}}^T \mathcal{R}_c^{-1}\\
      \hat{\mathbf{x}}^{\kappa + 1} &=  \hat{\mathbf{x}}^{\kappa} \!\boxplus\! (-\mathbf{K}_c \mathbf{h}_c(\hat{\mathbf{x}}^{\kappa}, \mathbf{0}) \!-\! (\mathbf{I}-\mathbf{K}_c\mathbf{H}_c^{\kappa}) (\hat{\mathbf{x}}^{\kappa} \boxminus \hat{\mathbf{x}}^0)).
    \end{aligned}
    \label{eq:leg_update}
  \end{equation}

  After convergence or the maximum iterations reached, the state $\bar{\mathbf{x}}$ is updated by kinematics with the covariance $\bar{\mathbf{P}}$ is:
  \begin{equation}
    \begin{aligned}
      \bar{\mathbf{x}} = \hat{\mathbf{x}}^{\kappa + 1}, \bar{\mathbf{P}} = (\mathbf{I} - \mathbf{K}_c \mathbf{H}_c)\hat{\mathbf{P}}.
    \end{aligned}
    \label{eq:leg_update_state}
  \end{equation}

\subsection{Sequential Phase}   %--------------------------------------------------SEQ
  % The sequential updates for LiDAR and camera have been proven in \cite{FAST-LIVO2} to be theoretically equivalent to standard updates while providing improved robustness and efficiency.
  % We adopt a similar strategy here, briefly introduced in this section.
  \subsubsection{LiDAR Measurement}
  When LiDAR and visual data are acquired, the LiDAR update is performed first. 
  A necessary step is to recombine the LiDAR scan to synchronize it with the image. 
  Additionally, distortion correction of points is performed here, rather than following state prediction.
  Instead of backward-propagating the predicted states, we perform this process by using a state buffer that maintains the updated states from the asynchronous phase.
  Then, the undistorted points ${ \{{}^L\mathbf{p}_j\} }$ of a subscan are obtained, and the point-to-plane measurement model for the $j$th point is given by:
  \begin{equation}
    \begin{aligned}
      \mathbf{0} \!=\! \mathbf{h}_l&(\mathbf{x}, \mathbf{n}_{l,j}) \triangleq ({}^W\!\mathbf{u}_j^{gt})^{\!T} ({}^W\!\mathbf{T}_{L} ({}^L\mathbf{p}_j + \mathbf{n}_{p,j}) - {}^W\!\mathbf{q}_j^{gt}) \\
                 &{}^W\mathbf{u}_j^{gt} = {}^W\mathbf{u}_j + \mathbf{n}_{u, j}, {}^W\mathbf{q}_j^{gt} = {}^W\mathbf{q}_j + \mathbf{n}_{q, j}
    \end{aligned}
    \label{eq:LiDAR_meas}
  \end{equation}
  where ${}^W\mathbf{T}_{L} = {}^W\mathbf{T}_{I} {}^I\mathbf{T}_L$ implicitly includes the state vector, $({}^W\mathbf{u}_j, {}^W\mathbf{q}_j)$ are the normal and the center point of the plane corresponding to ${}^L\mathbf{p}_j$, and $\mathbf{n}_l \sim \mathcal{N}(\mathbf{0}, \mathcal{R}_l) = (\mathbf{n}_u, \mathbf{n}_q, \mathbf{n}_p)$ denotes the noise associated with ${}^W\mathbf{u}_j$, ${}^W\mathbf{q}_j$, and ${}^L\mathbf{p}_j$, respectively.
  % , see \cite{FAST-LIVO2} for more details.

  \subsubsection{Visual Measurement}
  We follow \cite{FAST-LIVO2} to select the visual map points $\{ {}^W\mathbf{p}_j \}$ from the voxel map, and project them onto the current image. 
  The pixel errors between the current and reference patches then constitute the sparse-direct visual measurement:
  \begin{equation}
    \begin{aligned}
      \mathbf{0} = &\mathbf{h}_v(\mathbf{x}, \mathbf{n}_{v})
                 \triangleq \tau_i (\mathbf{I}_i( \boldsymbol{\pi}({}^C\mathbf{T}_W {}^W\mathbf{p}_j ) + \Delta \mathbf{u}) + \mathbf{n}_{v,i}) \\
                 &- \tau_r (\mathbf{I}_r( \boldsymbol{\pi}({}^{C_r}\mathbf{T}_W {}^W\mathbf{p}_j ) + \mathbf{A}_{j}^{r} \Delta \mathbf{u}) + \mathbf{n}_{v,r})
    \end{aligned}
    \label{eq:Visual_meas}
  \end{equation}
  where $\mathbf{I}_i(\cdot)$ and $\mathbf{I}_r(\cdot)$ are the current image at $t_i$ and the reference image, $\boldsymbol{\pi}(\cdot)$ is the camera projection model, $\Delta\mathbf{u}$ is the pixel offset within the current patch centered at the projected point, ${}^{C_r}\mathbf{T}_W$ is the transformation from the world frame to the reference frame, $\mathbf{A}_j^{r}$ is the affine warping matrix from the $j$th patch to the reference patch, and $\mathbf{n}_v \sim \mathcal{N}(\mathbf{0}, \mathcal{R}_v) = (\mathbf{n}_{v, i}, \mathbf{n}_{v, r})$ is the visual measurement noise.

  \subsubsection{Iterative Update}
  In the sequential phase, LiDAR and camera measurements are synced via scan recombination, and the state is updated by LiDAR first.
  Specifically, linearizing the measurement models (\ref{eq:LiDAR_meas}) at $\hat{\mathbf{x}}^{\kappa}$ leads to the residual $\mathbf{r}_l$:
  \begin{equation}
    \begin{aligned}
      \mathbf{r}_l =
      \mathbf{h}_l(\mathbf{x}, \mathbf{n}_l) - \mathbf{h}_l(\hat{\mathbf{x}}^{\kappa}, \mathbf{0})
      \simeq \mathbf{H}_l^{\kappa}\tilde{\mathbf{x}}^{\kappa} + \mathbf{D}_l^{\kappa} \mathbf{n}_l
    \end{aligned}
    \label{eq:LiDAR_Res}
  \end{equation}
  where $\mathbf{H}_l^{\kappa}$ and $\mathbf{D}_l^{\kappa}$ are the Jacobian matrices of residual $\mathbf{r}_l$ with respect to $\tilde{\mathbf{x}}^{\kappa}$ and $\mathbf{n}_l$, respectively.

  Based on the ESIKF, LiDAR update calculates the Kalman gain $\mathbf{K}_l$ and iteratively updates the state as:
  \begin{equation}
    \begin{aligned}
      \mathbf{K}_l &= ({\mathbf{H}_l^{\kappa}}^T \mathcal{R}_l^{-1} \mathbf{H}_l^{\kappa} + \hat{\mathbf{P}}^{-1})^{-1} {\mathbf{H}_l^{\kappa}}^T \mathcal{R}_l^{-1}\\
      \hat{\mathbf{x}}^{\kappa + 1} &= \hat{\mathbf{x}}^{\kappa} \! \boxplus \! (-\mathbf{K}_l \mathbf{h}_l(\hat{\mathbf{x}}^{\kappa}, \mathbf{0}) \!-\! (\mathbf{I}-\mathbf{K}_l\mathbf{H}_l^{\kappa}) (\hat{\mathbf{x}}^{\kappa} \boxminus \hat{\mathbf{x}}^0)).
    \end{aligned}
    \label{eq:seq_update}
  \end{equation}

  After convergence or the maximum iterations reached, the LiDAR-updated state $\bar{\mathbf{x}}$ with the covariance $\bar{\mathbf{P}}$ is given by:
  \begin{equation}
    \begin{aligned}
      \bar{\mathbf{x}} = \hat{\mathbf{x}}^{\kappa + 1}, \bar{\mathbf{P}} = (\mathbf{I} - \mathbf{K}_l \mathbf{H}_l)\hat{\mathbf{P}}
    \end{aligned}
    \label{eq:seq_update_state}
  \end{equation}
  where $\bar{\mathbf{x}}$ and $\bar{\mathbf{P}}$ are then used as the prior for the visual update, which follows the same update form with $(\mathbf{h}_v, \mathbf{H}_v, \mathcal{R}_v)$, thus the derivation is omitted for brevity.

\subsection{Multimodal Adaptation} \label{sec:4-e}
To fully utilize the advantages of sensor fusion, KILVO adaptively and seamlessly transitions to feasible modalities {according to the health status of the} measurement-level sensors, {thereby achieving} strong robustness against sensor failures {and degradation}.
{The health status is defined by data-stream integrity and measurement reliability.}
{For data-stream integrity,} we monitor a global time $t_{chk}$ shared across the sensors, and the last-update time $t_{end, s}$ of each sensor $s$.
When $\Delta t_{chk, s}$ exceeds a threshold $\theta_s$, the corresponding sensor is considered lost, and the modality will be switched:
\begin{equation} 
  \begin{aligned}
    \Delta t_{chk, s} = t_{chk} - t_{end, s} > \theta_s
  \end{aligned}
\end{equation}
where $\theta_s$ is set {based on the sensor frequency, e.g.,} 0.05 s for 1 kHz encoders and 0.3 s for 10 Hz LiDAR/camera.
{
For measurement reliability, KILVO can flexibly integrate degradation indicators to trigger the same modality transition.
In our work, LiDAR degradation is evaluated from the spatial distribution of points, while visual degradation is detected according to the number of valid visual map points.
}

{Each modality of KILVO is} shown in Fig. \ref{fig_2}.
{Once the active modality changes, the data synchronization strategy, measurement update, buffer management, and contact estimation are reconfigured consistently in the unified estimator.}
When the image/LiDAR is {unavailable}, the system is driven by the asynchronous phase.
For example, in KIL modality, the states are updated and output at 1 kHz, and mapping with geometry rather than colored points.
In KI modality, the historical ${}^W \! \mathbf{p}_c$ {are cached instead of} the map points to fit ground patches and compute the foot clearance $c_i$, thereby maintaining contact estimation {without relying on LiDAR points}.
Notably, LiDAR failure will trigger the KI modality even with valid images, because visual updates rely on the map's geometry.
Once the leg kinematics is {unavailable}, the system is {transitioned} to LIV/LI modality, driven in sequential phase while contact estimation is suspended.
In addition, the state buffer, which holds the updated states, will be cleared, and then the distortion correction of points is performed based on the predicted states from IMU backward propagation instead.

For a non-full modality, {the health status of} all sensors is still checked, allowing KILVO to {recover} to a richer modality when the measurements are available again.
To avoid introducing low-quality measurements during ground impacts, we transition back only when the acceleration normalized deviation $d_a$ remains below $\theta_d$ (details in Section \ref{sec:5-b}).
\section{Contact Estimation} \label{sec:5}
  % For compact and efficient deployment, we design a contact estimation module tightly in KILVO.
  % In this section, we describe how to utilize only the extisting information to deliver stable contact estimation with low computational load.
\subsection{Dynamic Initialization} \label{sec:5-a}
  The main objective of initialization is to compute the swing-reference statistics associated with the {body} acceleration, foot velocity {of the supported leg}, and kinematic residual, denoted as $\mathcal{S}^{sw} = ((\boldsymbol{\mu}_a^{sw}, \boldsymbol{\sigma}_a^{sw}), (\boldsymbol{\mu}_{fv}^{sw}, \boldsymbol{\sigma}_{fv}^{sw}), (\boldsymbol{\mu}_r^{sw}, \boldsymbol{\sigma}_r^{sw}))$.
  The acceleration and foot velocity statistics are used for contact/non-contact detection, and the residual statistics are used for kinematic noise adaptation (see Section \ref{sec:4-c}).

\begin{figure}[!t]
\centering
\includegraphics[width=\linewidth]{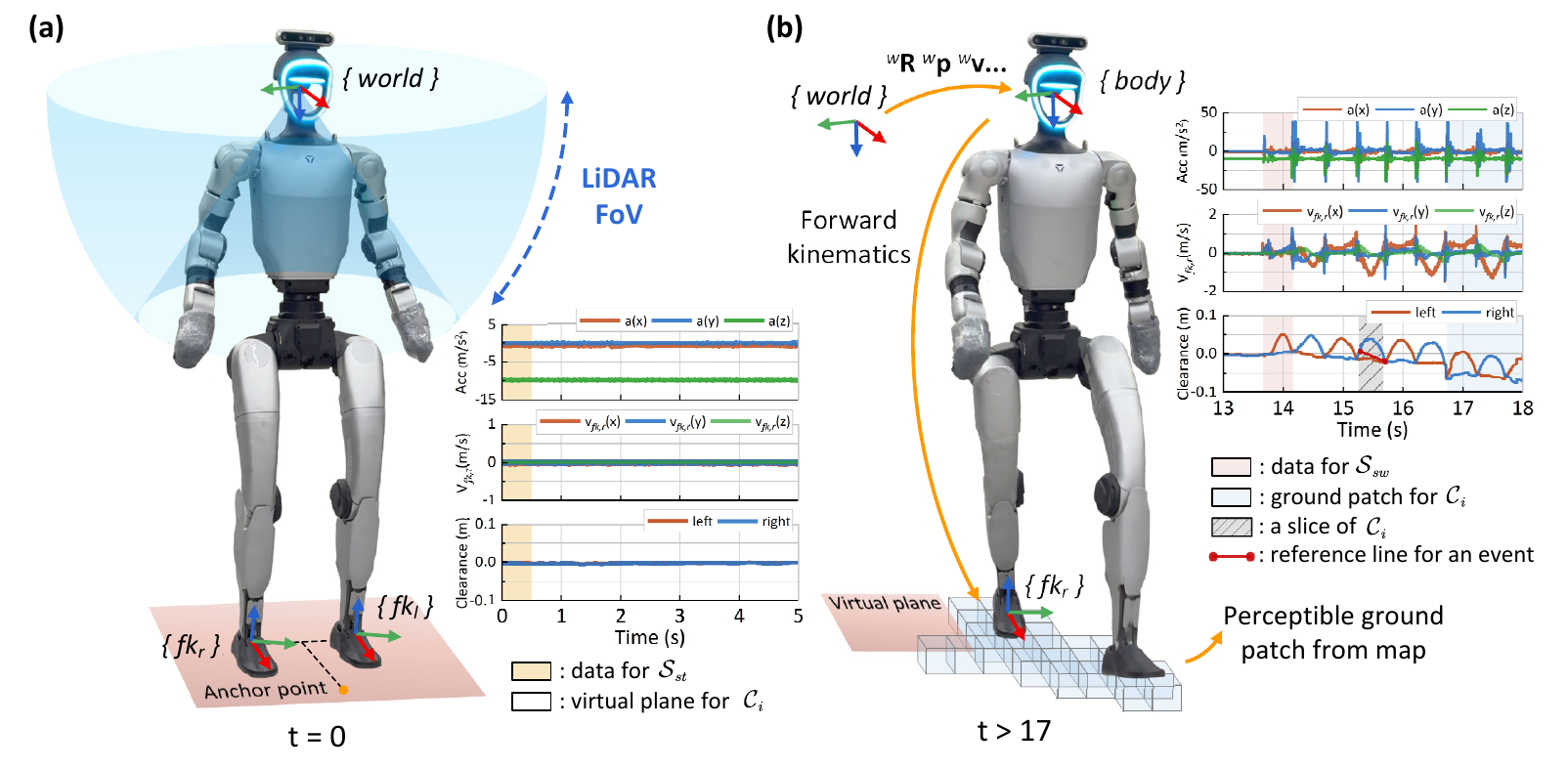}
\caption{Illustration of contact estimation. (a) Dynamic initialization after system startup, until the first step is completed. (b) Contact estimation based on $\mathcal{S}^{{sw}}$ and the perceptible ground.}
\label{fig_3}
\end{figure}

  To obtain $\mathcal{S}^{sw}$, we first compute the static reference statistics $\mathcal{S}^{st}$,
  which allows estimating contacts even during initialization.
  As shown in Fig. \ref{fig_3}(a), assuming both feet are in contact at the system start, the data are recorded during an initial stationary period for $\mathcal{S}^{st}$.
  It is used to compute the frame-wise deviations to detect the first lift-off and touchdown, thereby obtaining the swing cycle $\Delta t_{sw} \triangleq t_{td} - t_{lo}$.
  For each signal $\mathbf{z} \in \{\mathbf{a}, \mathbf{v}_{fk}, \mathbf{r}_c\}$, $\mathcal{S}^{sw}$ is calculated as follows: 
  \begin{equation}
    \begin{aligned}
      \boldsymbol{\mu}^{sw}_{z} \!=\! \frac{1}{N_g} \! \sum_{i=1}^{N_g} \! \mathbf{z}(t_i), (\boldsymbol{\sigma}^{sw}_z)^2 \!=\! \frac{1}{N_g} \! \sum_{i=1}^{N_g} \! (\mathbf{z}(t_i) - \boldsymbol{\mu}_z^{sw}) ^ 2
    \end{aligned}
  \end{equation}
  where $t_i \in [t_{lo}, t_{td}]$ and the square is applied element-wise.

  Moreover, a sliding window $\mathcal{C}$ of foot clearances is required for contact detection. 
  At each contact moment $t_j$ with foot position ${}^W\!\mathbf{p}_{fk_j}$, we locate the root or subvoxel where ${}^W\!\mathbf{p}_{fk_j}$ lies in the map to obtain the ground patch $({}^W\!\mathbf{u}_j, {}^W\!\mathbf{q}_j)$. 
  Using this patch, which is reused until the next contact update, we compute the current clearance $c_i$ and save it in $\mathcal{C}_i$: 
  \begin{equation} 
    \begin{aligned} 
      c_i \!=\! {}^W\!\mathbf{u}_j^T( {}^W\!\mathbf{T}_I {}^I\mathbf{p}_{fk, i} \!-\! {}^W \! \mathbf{q}_j), 
      \mathcal{C}_i = \{c_{k}\}_{t_k \in [t_i - \Delta t_{sw}, t_i]}.
    \end{aligned}
  \end{equation}
  And notice that the LiDAR blind zones often leave no voxels during initialization, the ground patch is instead fitted by the stance feet and an expansion anchor point. 
  We {also} cache at least three contact foot positions for ground patch fitting in KI modality.
  {The above designs together reduce the dependence of} the contact estimation module on the {point cloud map}.

\subsection{Contact/Non-contact Detection} \label{sec:5-b}
  The {contact state of each foot is detected separately} by complementarily leveraging the {reference statistics} $(\boldsymbol{\mu}_a^{sw}, \boldsymbol{\sigma}_a^{sw})$ and $(\boldsymbol{\mu}_{fv}^{sw}, \boldsymbol{\sigma}_{fv}^{sw})$.
  {Unlike sliding-window-based acceleration variance used in \cite{HR2-KILO}, we} calculate {frame-wise} normalized deviations regarding acceleration and foot velocity, denoted as $d_a$ and $d_{fv}$:
  \begin{equation}
    \begin{aligned}
      {d}_{a} \!=\! \boldsymbol{\alpha}_a^T \! \left| \frac{({\mathbf{a}_{m}} \!-\! \mathbf{b}_{a}) \!-\! \boldsymbol{\mu}_a^{sw}}{\boldsymbol{\sigma}_a^{sw}} \right|,
      {d}_{fv} \!=\! \boldsymbol{\alpha}_{fv}^T \! \left| \frac{{}^W\mathbf{v}_{{fk}} \!-\! \boldsymbol{\mu}_{fv}^{sw}}{\boldsymbol{\sigma}_{fv}^{sw}} \right|
    \end{aligned}
  \end{equation}
  where $\boldsymbol{\alpha}_a$ and $\boldsymbol{\alpha}_{fv}$ are the weight vectors, set as unit vectors,
  $\mathcal{S}^{sw} = \mathcal{S}^{st}$ for the first contact.
  Let $(\theta_{d}, N_d)$ denote the detection thresholds (set to 10 and 5 in this article), which are robust across terrains and gaits due to the dimensionless normalization.
  {When $d_a$ or $d_{fv}$ remains greater than $\theta_d$ for $N_d$ frames, the corresponding cue is considered activated.}
  
  For a non-contact foot, when {acceleration cue is activated}, we take the slice from sliding window $\mathcal{C}_i$, and compute the signed area between the clearance trajectory and its respective reference lines, as shown in Fig. \ref{fig_3}(b).
  We then identify the foot with the larger area as having just completed a swing and set it to contact.
  In addition, when $d_{fv} < \theta_d$ lasts for $N_d$ frames, we use a shorter slice of $\mathcal{C}_i$ to estimate the recent clearance trend $k_{c}$ by linear regression.
  If {$k_c$ indicates} a descending foot-clearance trend, the foot is also set to contact.
  Similarly, for non-contact detection,
  {a contact} foot will be treated as non-contact if {the corresponding foot velocity cue is activated}.

  Notice that ${d}_{a}$ and ${d}_{fv}$ tend to fluctuate in a coupled manner in practical contact events.
  To minimize false detection, when a foot is set to contact, it is allowed to be non-contact only after an acceleration settling period, i.e., $d_a < \theta_d$ for $N_d$ frames.
  Conversely, a non-contact foot needs to experience a swing peak ($d_{fv} < \theta_d$) before it switches to contact.
\section{Experiments}
\subsection{Implementation \& Datasets}
  \begin{figure}[t]
  \centering
  \includegraphics[width=\linewidth]{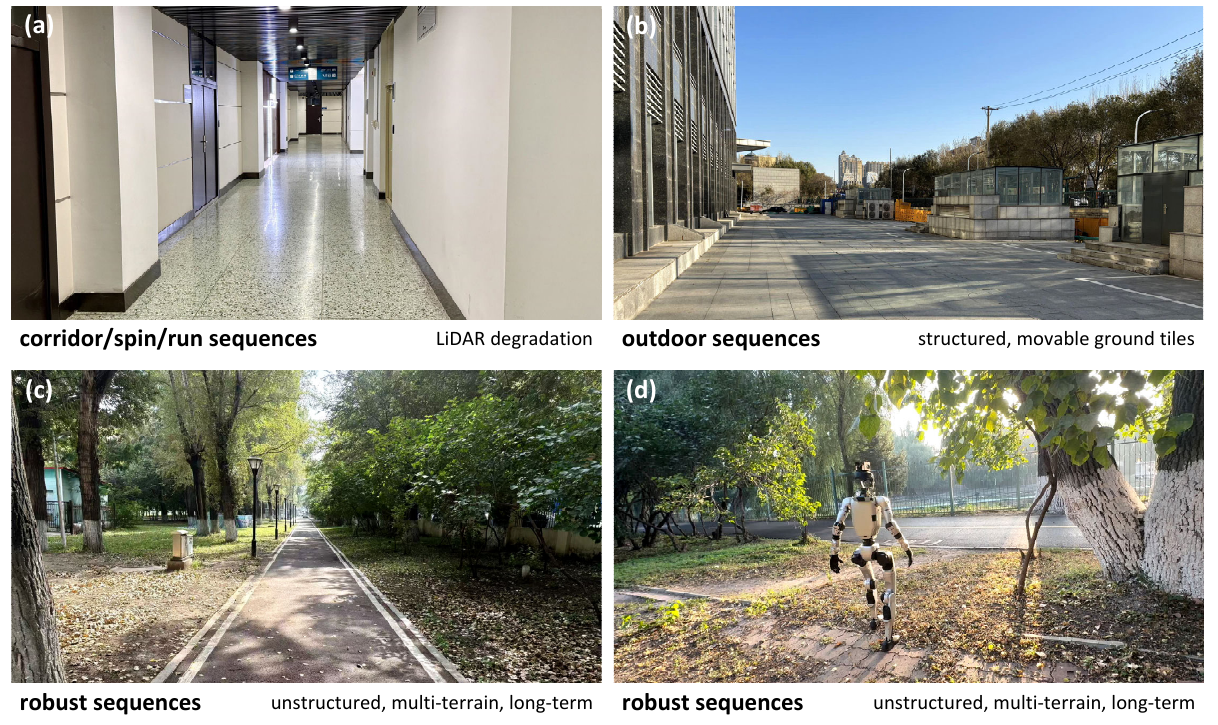}
  \caption{Environments of the real-world experiments. (a) presents an indoor corridor with LiDAR degradation. (b) is the outdoor environment with movable ground tiles. (c) and (d) are the unstructured outdoor environments, including asphalt roads, rubber roads, brick pavements, and dirt paths.}
  \label{fig_dataset}
  \end{figure}

  We implement KILVO in C++ and ROS. 
  The leg kinematic calculation is based on Pinocchio \cite{pinocchio}.
  The sensor extrinsics are from algorithm calibration or manufacturer specifications.
  Extensive experiments are conducted on public datasets and in the real world.
  We first validate the contact estimation module, and then evaluate the overall performance in accuracy, efficiency, and robustness against SOTA fusion methods.
  For fair comparison, all the methods share the same common parameters, and the remaining are set according to their default configurations.
  The computational platform is a desktop (Intel i5-13490F CPU and 16 GB RAM), with data collection deployed on a Jetson Orin NX (Arm Cortex-A78 and 16 GB RAM) embedded within the robot.

  The public humanoid robot datasets are from LIKO \cite{LIKO} and HR$^2$-KILO \cite{HR2-KILO}. 
  LIKO dataset is collected in a room by a BHR-B3 robot equipped with a Velodyne VLP16 LiDAR (360$^{\circ}$$\times$30$^{\circ}$ FoV, 10 Hz), an Xsens Mti-100 IMU (200 Hz), HDH EBI-1135 Encoders (1 kHz), and SRI F/T sensors (1 kHz).
  The ground truth is provided by a VICON motion capture system. 
  HR$^2$-KILO employs a Livox Mid360 (360$^{\circ}$$\times$59$^{\circ}$ FoV, 10 Hz) with a built-in ICM40609 IMU (200 Hz), and Unitree self-developed encoders (1 kHz).
  It collects different gaits of the humanoid robot in rooms and corridors.

  Since the above datasets exclude cameras and are limited to indoor environments, more comprehensive experiments are conducted in real world, 
  which comprise 15 sequences across different sensor combinations, multiple gaits, scenarios, and sensor failure events.
  The specific environments can be seen in Fig \ref{fig_dataset}.
  \begin{figure}[t]
    \centering
    \includegraphics[width=\linewidth]{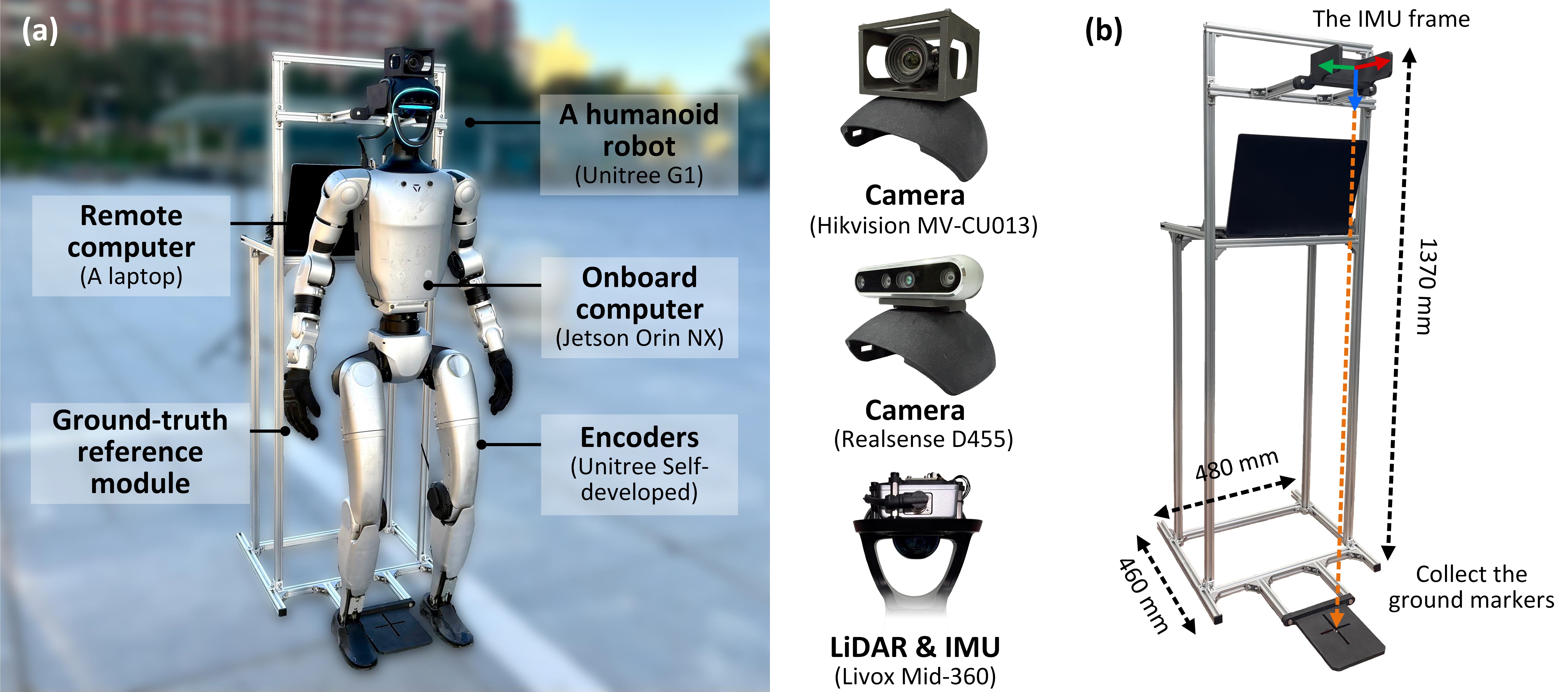}
    \caption{The humanoid robot with the equipment used for the real-world experiments. (a) Robot platform and sensor setup. (b) Mechanical module for recording relative ground truth.}
    \label{fig_roboconfig}
  \end{figure}  
  \begin{table}[t!]  %--------------------------------------------------------- TABLE BEG
    \caption{Estimated Contact Event Results (Left/Right)} \label{tab:contact1}
    \centering
    \begin{threeparttable}
      \begin{tabularx}{\linewidth}{p{1.4cm} *{5}{>{\centering\arraybackslash}X}}
        \toprule
        {Sequences}        & {GRF} & {DCE} & {HRC} & {Ours} & {GT}\\
        \midrule
        {fwd\_bwd}        & --\tnote{a} & -- & 74/73   & 75/75 & 75/76 \\
        {square\_walk}    & -- & -- & 86/82   & 87/86 & 87/88 \\
        {square\_walk\_l} & -- & -- & 134/134 & 164/165 & 166/165 \\
        {walk\_in\_place} & -- & -- & 202/202 & 203/204 & 203/204 \\
        {up\_slope}       & -- & -- & 73/71   & 72/74 & 74/75 \\
        \midrule
        {walk\_back}      & -- & 14/11     & 14/13    & 16/16 & 16/16 \\
        {fwd\_run}        & 8/7         & 7/5       & 7/6    & 9/8 & 9/8  \\
        {soft\_terrain}   & 9/9         & 7/7       & 7/7    & 9/10 & 9/10 \\
        \bottomrule
      \end{tabularx}
        \begin{tablenotes}[para,flushleft]
          \footnotesize \item[a] -- denotes that the sequences do not produce the joint position, velocity or torque data, which are necessary for the corresponding algorithms.
        \end{tablenotes}
      \end{threeparttable}
  \end{table} %---------------------------------------------------------------- TABLE END
  These sequences are collected by a Unitree G1 with a customized sensor suite [Fig. \ref{fig_roboconfig} (a)], including a Hikvision MV-CA013-A0UC camera (54.3$^{\circ}$$\times$44.6$^{\circ}$ FoV, 10 Hz), a Realsense D455 camera (90$^{\circ}$$\times$65$^{\circ}$ FoV, 10 Hz), a Livox Mid360 LiDAR (360$^{\circ}$$\times$59$^{\circ}$ FoV, 10 Hz) with the built-in IMU (200 Hz), and the Unitree self-developed encoders (1 kHz).
  The sequences labeled ``h" and ``r" correspond to image data from the Hikvision and Realsense cameras, respectively.
  For quantitative evaluation, a mechanical module is designed to record the ground truth of the humanoid robot [Fig. \ref{fig_roboconfig} (b)].
\subsection{Contact Estimation}
\begin{figure}[!t]  %-------------------------------------------------------- FIGUR BEG
    \centering
    \includegraphics[width=0.95\linewidth]{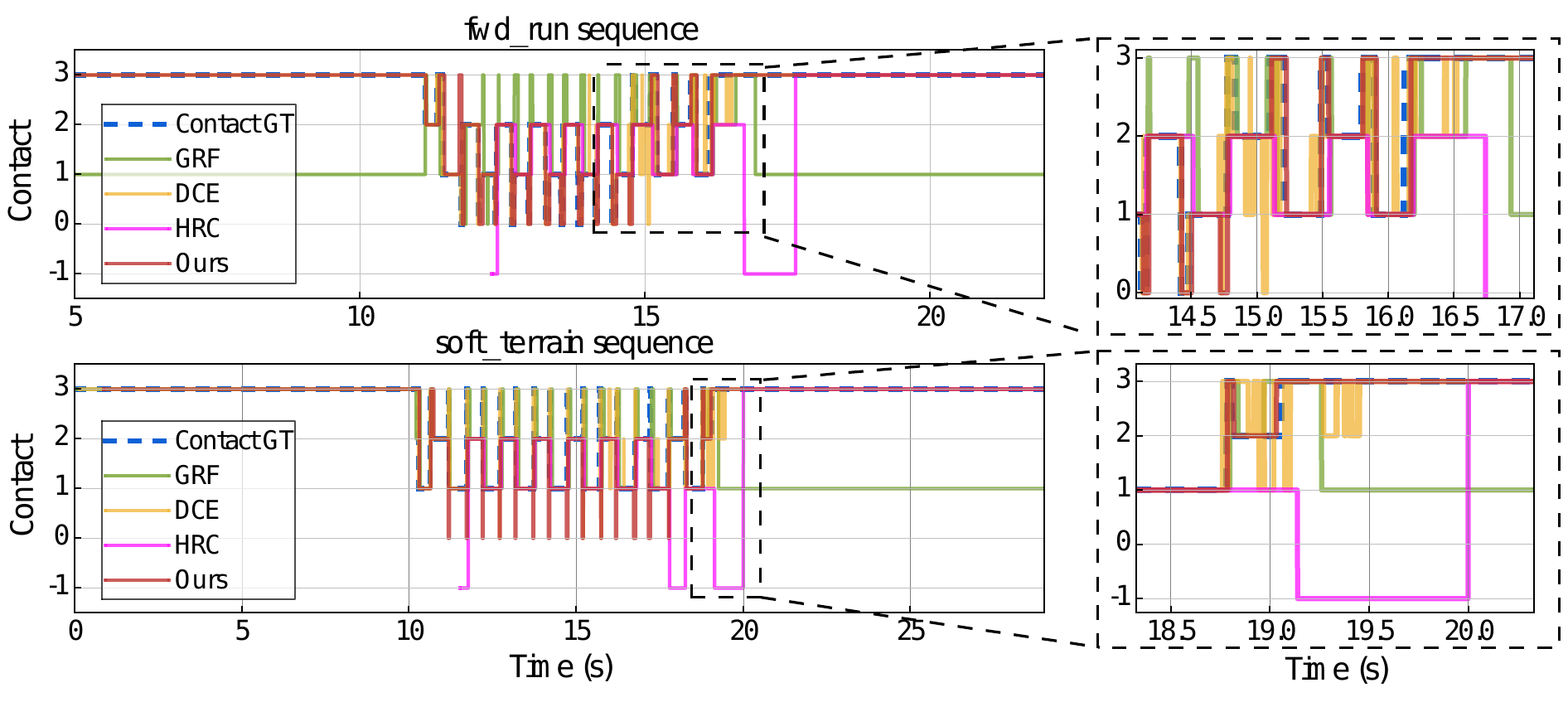}
    \caption{Contact estimation results on fwd\_run and soft\_terrain sequences.}
    \label{fig_contact_acc}
  \end{figure}

  In this section, we evaluate the contact estimation module based on the public datasets.
  The comparisons include thresholding GRF (calculated as described in \cite{contact_TO_2017, contact_ProEsti_2020}), the learning-based method (denoted as DCE) \cite{contact_learn_2021}, and the contact detection module of HR$^2$-KILO\cite{HR2-KILO} (denoted as HRC).
  The accuracy of detected contact events is shown in Table \ref{tab:contact1}.
  Since misdetections are more severe in mid-contact than at transitions, any error within a contact phase is counted as an incorrect event.
  KILVO produces results that are closest to the ground truth on most sequences, exhibiting satisfactory detection stability.
  We further report the accuracy and false positive rate (FPR) in Table \ref{tab:contact2}.
  Our method achieves the highest accuracy and lowest FPR in 7 of 8 sequences. For all the sequences, we maintain over 95\% accuracy while holding an average FPR of 3.67\%.
  HRC is limited by its inability to detect foot uncontact, {time-window-based} estimation delays, and {missing} the first two contacts due to initialization.
  DCE exhibits degradation on ``fwd\_run'' and ``soft\_terrain'' (detailed in Fig. \ref{fig_contact_acc}) given that fully training the model across all scenarios is impractical.
  DCE suffers accuracy loss in the latter of ``fwd\_run'' because the robot experiences an emergency stop, which is not covered in our training set.
  For ``soft\_terrain'', the final slight step is successfully detected only by ours and GRF.
  However, GRF is highly threshold-dependent and fails after 19 s, as the robot stops on the soft terrain, causing CoM displacement and uneven force distribution.
  \begin{table}[tb] %--------------------------------------------------------- TABLE BEG
    \caption{Accuracy and False Positive Rate of Contact Estimation} \label{tab:contact2}
    \begin{center}
      \begin{threeparttable}
        \begin{tabularx}{\linewidth}
        {>{\arraybackslash}p{1.3cm} *{4}{>{\centering\arraybackslash}X} | *{4}{>{\centering\arraybackslash}X}}
        \toprule % 第1道横线
        \multirow{2}{*}{Sequences} & \multicolumn{4}{c|}{Leg Avg Accuracy (\%)} & \multicolumn{4}{c}{Leg Avg FPR (\%)} \\
        \cmidrule(lr){2-5} \cmidrule(lr){6-9}
                            & GRF    & DCE   & HRC   & Ours  & GRF   & DCE   & HRC   & Ours\\
        \midrule % 第2道横线
          {fwd\_bwd}	       &   --\tnote{a}  &   --  & 90.47 & 97.42 &   --  &   --  &  9.56 & 2.00\\
          {square\_walk} 		 &   --  &   --  & 87.61 & 95.04 &   --  &   --  & 11.86 & 5.15\\
          {square\_walk\_l}	 &   --  &   --  & 70.84 & 96.92 &   --  &   --  & 22.41 & 2.06\\
          {walk\_in\_place}  &   --  &   --  & 91.24 & 96.91 &   --  &   --  &  8.82 & 0.64\\
          {up\_slope}			   &   --  &   --  & 93.42 & 97.14 &   --  &   --  &  9.91 & 6.94\\
          \midrule % 第3道横线
          {walk\_back} 		   &   --  & 99.35 & 93.27 & 98.21 &   --  & 1.84  &  7.69 & 5.50\\
          {fwd\_run}	       & 58.54 & 97.69 & 90.77 & 98.44 & 21.70 & 5.53  & 14.52 & 4.88\\
          {soft\_terrain}    & 79.69 & 97.89 & 95.74 & 98.33 & 15.08 & 6.23  &  8.00 & 2.15\\
        \bottomrule % 第4道横线
        \end{tabularx}
        \begin{tablenotes}[para,flushleft]
          \footnotesize \item[a] -- denotes that the sequences do not produce the joint position, velocity or torque data, which are necessary for the corresponding algorithms.
        \end{tablenotes}
      \end{threeparttable}
    \end{center}
  \end{table} %---------------------------------------------------------------- TABLE END

          %-------------------------------------------------------- FIGUR END
  
\begin{figure}[!t]  %-------------------------------------------------------- FIGUR BEG
    \centering
    \includegraphics[width=\linewidth]{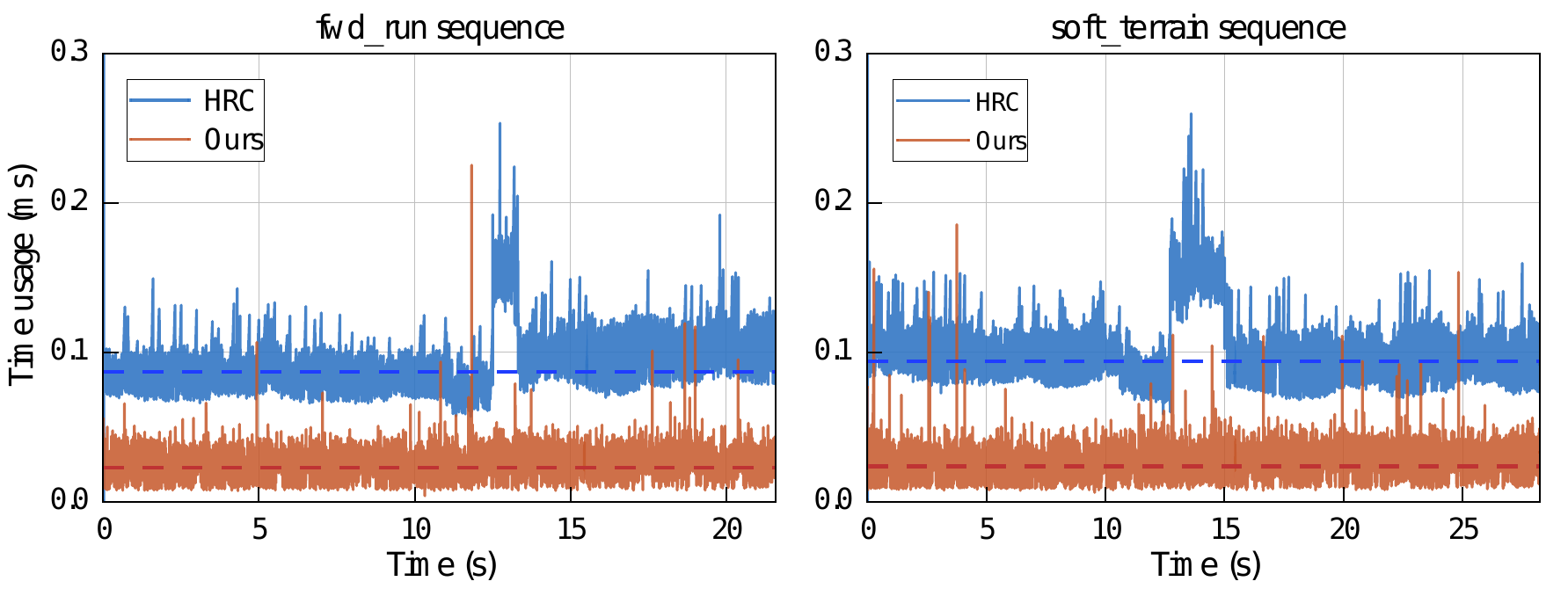}
    \caption{Time usage for contact estimation of HR$^2$-KILO and KILVO on fwd\_run and soft\_terrain sequences.}
    \label{fig_contact_eff}
  \end{figure}

  Then, we analyze the computational efficiency of this module.
  The time consumption comparison between our method and HRC is shown in Fig. \ref{fig_contact_eff}.
  On ``fwd\_run'' and ``soft\_terrain'', the average processing time of HRC is 0.08 ms and 0.09 ms, respectively.
  KILVO achieves a 76\% efficiency improvement with an average cost of about 0.02 ms.
  This benefits from our method that retrieves map points and builds the ground patch only at foot-contact moments, which is reused within a contact event rather than being processed frame by frame.
          %-------------------------------------------------------- FIGUR END

  \begin{table}[t!]  %--------------------------------------------------------------- TABLE BEG
    \caption{Benchmark Results on Public Datasets} \label{tab:accuracy1}
    \centering
    \begin{threeparttable}
      \begin{tabularx}{\linewidth}{p{1.45cm} *{6}{>{\centering\arraybackslash}X}}
        \toprule
        {Sequence} &      {LIO-SAM}&{FAST-LIO2}&{FAST-LIVO2*}&{LIKO/ LIKO*}&{HR$^2$-KILO}&KILVO (KIL)\\
        \midrule
        \multicolumn{2}{l}{LIKO Dataset} & \multicolumn{5}{c}{ATE RMSE (m)}\\
        \midrule
        {fwd\_bwd}        & 0.0186          & 0.0129 & 0.0132             & 0.0127 & \textbf{0.0071} & \underline{0.0080} \\
        {square\_walk}    & 0.0207          & 0.0270 & \underline{0.0103} & 0.0280 & 0.0106          & \textbf{0.0100} \\
        {square\_walk\_l} & 0.0232          & 0.0175 & \textbf{0.0139}    & 0.0182 & 0.0140          & \textbf{0.0139} \\
        {walk\_in\_place} & \textbf{0.0164} & 0.0275 & 0.0255             & 0.0293 & 0.0244          & \underline{0.0234} \\
        {up\_slope}       & 1.0960          & 0.0327 & \underline{0.0235} & 0.0417 & 0.0260          & \textbf{0.0204} \\
        \midrule
        \multicolumn{2}{l}{LIKO Dataset} & \multicolumn{5}{c}{RTE RMSE (\%)}\\
        \midrule
        {fwd\_bwd}        & 0.0028 & 0.0064 & 0.0080 & 0.0014             & \underline{0.0013} & \textbf{0.0011} \\
        {square\_walk}    & 0.0027 & 0.0082 & 0.0059 & 0.0017             & \underline{0.0013} & \textbf{0.0010} \\
        {square\_walk\_l} & 0.0029 & 0.0079 & 0.0069 & 0.0019             & \underline{0.0017} & \textbf{0.0011} \\
        {walk\_in\_place} & 0.0026 & 0.0072 & 0.0049 & 0.0018             & \underline{0.0013} & \textbf{0.0009} \\
        {up\_slope}       & 0.0112 & 0.0097 & 0.0117 & \underline{0.0015} & 0.0016             & \textbf{0.0008} \\
        \midrule
        \multicolumn{2}{l}{HR$^2$-KILO Dataset} & \multicolumn{5}{c}{End-to-end Error (m) on the Z-axis}\\
        \midrule
        {long\_corr}      &$\times$\tnote{a}& 0.0178 & \underline{0.0060} & 0.0076 & 0.0062 & \textbf{0.0027} \\
        {short\_corr}     & 0.0140          & 0.0093 & 0.0020 & 0.0096 & \textbf{0.0012} & \underline{0.0016} \\
        {switch\_run}     & 0.0098          & 0.0251 & 0.0097 & 0.0128 & \textbf{0.0019} & \underline{0.0061} \\
        {spinning}        & 0.0108          & 0.0063 & \underline{0.0023} & 0.0068 & 0.0027 & \textbf{0.0017} \\
        \bottomrule
      \end{tabularx}
    \end{threeparttable}
      \begin{tablenotes}[para,flushleft]  %不分点+左对齐
          \footnotesize \item[a] $\times$ denotes that the system failed.
      \end{tablenotes}
  \end{table} %---------------------------------------------------------------------- TABLE END

\subsection{Accuracy Evaluation}
  \subsubsection{Public Dataset Benchmark}
    We first evaluate KILVO against LiDAR-inertial(-kinematic) methods on the public datasets, including LIO-SAM \cite{LIO-SAM} (without loop closure), FAST-LIO2 \cite{FAST-LIO2}, the LIO configuration of FAST-LIVO2 \cite{FAST-LIVO2} (denoted as FAST-LIVO2*), LIKO \cite{LIKO}, and HR$^2$-KILO \cite{HR2-KILO}. 
    For LIKO dataset, the absolute translation errors (ATE) and relative translation errors (RTE) are computed by EVO \cite{evo}.
    The end-to-end errors on the Z-axis are reported for HR$^2$-KILO dataset.
    Additionally, the Unitree G1 is not equipped with contact sensors, leading to LIKO operation without kinematic constraints in some sequences, denoted as LIKO*.

    The benchmark results are shown in Table \ref{tab:accuracy1}. 
    On LIKO dataset, our approach achieves the lowest ATE in 3 of 5 sequences with an average RMSE of 0.0151 m.
    LIO-SAM and HR$^2$-KILO perform better on the remaining sequences, but our system remains competitive.
    In terms of RTE, KILVO outperforms others on all the sequences.
    For HR$^2$-KILO dataset, we maintain the end-to-end Z-axis errors below 1 cm, with optimal results in the sequences of ``long\_corr" and ``spinning".
    It is worth noting that KILVO, operating as its KIL modality on these datasets, delivers the best/second-best accuracy across the benchmarks.

  \subsubsection{Real-world Experiments}
    \begin{table*}[ht!] %--------------------------------------------------------- TABLE BEG
    \caption{End-to-end Translation Error (Meters) in Real World} \label{tab:Xaccuracy}
    \centering
    \begin{threeparttable}
      \begin{tabularx}{\linewidth}{p{1.5cm} *{11}{>{\centering\arraybackslash}X}}
        \toprule
        {Sequence} & 
        {LIO-SAM} & {FAST-LIO2} & 
        {R$^3$LIVE} & {FAST-LIVO2} & 
        {LIKO*} & {HR$^2$-KILO} & 
        {KILVO (LI)}   & {KILVO (LIV)} & {KILVO (KIL)} & KILVO & Duration (min:sec)\\
        \midrule            %LIO-SAM %FAST-LIO2           %R3LIVE           %FAST-LIVO2  %LIKO  % HR2KILO             %KILVO(LI)            %KILVO (LIV)         %KILVO (LIK)       %KILVO
        {corridor\_r01}&    $\times$\tnote{a} & 0.0076 &0.0316 &          0.0077 &     0.0037 & 0.0047             & 0.0062              & 0.0054             & \textbf{0.0018}  & \underline{0.0031}  &06 : 11\\
        {spin\_r01}    &    0.0152   & \underline{0.0040}&0.0144 &          0.0112 &     0.0163 & 0.0045             & 0.0113              & 0.0108             & 0.0144           & \textbf{0.0030}     &01 : 14\\
        {outdoor\_r01} &    $\times$ & 0.0206 &           0.2880 &          0.0248 &     0.0151 & 0.0134             & 0.0151              & 0.0162             & \textbf{0.0103}  & \textbf{0.0103}     &03 : 40\\
        {outdoor\_r02} &    0.5949   & 0.0311 &           0.0180 &          0.0064 &     0.0466 & 0.0103             & \underline{0.0031}  & 0.0073             & 0.0060           & \textbf{0.0005}     &02 : 49\\
        {outdoor\_r03} &    0.0076   & 0.0147 &           0.0863 &          0.0084 &     0.0191 & \textbf{0.0023}    & 0.0071              & 0.0060             & 0.0141           & \underline{0.0057}  &00 : 29\\
        \midrule
        {corridor\_h01}&    4.8363   & 0.0881 &           0.0884 &          0.0134 &     0.0818 & 0.0106             & 0.0015              & 0.0033             & \underline{0.0015}&\textbf{0.0005}     & 06 : 06\\
        {spin\_h01}    &    0.4279   & 0.0224 &           0.0152 &          0.0424 &     0.0203 & 0.0177             & \textbf{0.0139}     & \underline{0.0141} & 0.0169           & 0.0160              & 01 : 19\\
        {outdoor\_h01} &    $\times$ & 0.0273 &           0.0232 &          {\textbf{0.0116}} &     0.0328 & 0.0150             & 0.0150              & 0.0147 & 0.0242           & \underline{0.0125}     & 04 : 04\\
        {outdoor\_h02} &    1.8951   & \underline{0.0374}&\textbf{0.0195} & {0.0601} &     0.0401 & 0.0789             & 0.0703              & 0.0704             & 0.0711           & 0.0673              & 03 : 52\\
        {outdoor\_h03} &    $\times$ & 0.0708 &           0.0438 &          {0.0314} & 0.0424     & \underline{0.0145} & 0.0171              & 0.0171             & 0.0180           & \textbf{0.0120}     & 02 : 56\\
        {outdoor\_h04} &    0.0060   & 0.0102 &           0.0394 &          {0.0122} & 0.0161     & 0.0060             & \underline{0.0036}  & 0.0132             & \textbf{0.0015}  & 0.0054              & 01 : 02\\
        {run\_h01}     &    $\times$ & \underline{0.0197}&0.1207 &          0.0461 & 0.0402     & 0.0310             & \textbf{0.0157}     & 0.0203             & 0.0224           & 0.0223              & 00 : 40\\
        {run\_h02}     &    $\times$ & 0.0503 &           0.8018 &          0.0380 & 0.0457     & 0.0461             & 0.0476              & 0.0492             & \underline{0.0379}&\textbf{0.0373}     & 00 : 44\\
        % \midrule
        {robust\_h01}  &    $\times$ & 1.4520 &           4.9433 &          0.0242 & 1.9927     & 1.6309             & 0.2423              & \underline{0.0170} & 0.0235           & \textbf{0.0130}     & 09 : 14\\
        {robust\_h02}  &    $\times$ & 1.2003 &           0.6600 &          0.0237 & 0.9939     & 0.4141             & 0.0136              & 0.0129             & \underline{0.0110}&\textbf{0.0093}     & 06 : 15\\
        \midrule
        {Average}      &    1.1119   & 0.2038 &           0.4796 &          {0.0241} & 0.2271     & 0.1533             & 0.0322              & 0.0185             & \underline{0.0183} & \textbf{0.0145}  & 03 : 22\\
        \bottomrule
      \end{tabularx}
      \begin{tablenotes}[para,flushleft]  %不分点+左对齐
          \footnotesize \item[a] $\times$ denotes that the system failed.
      \end{tablenotes}
    \end{threeparttable}
  \end{table*} %---------------------------------------------------------------- TABLE END

  \begin{figure}[!t]
      \centering
      \includegraphics[height=3.8cm]{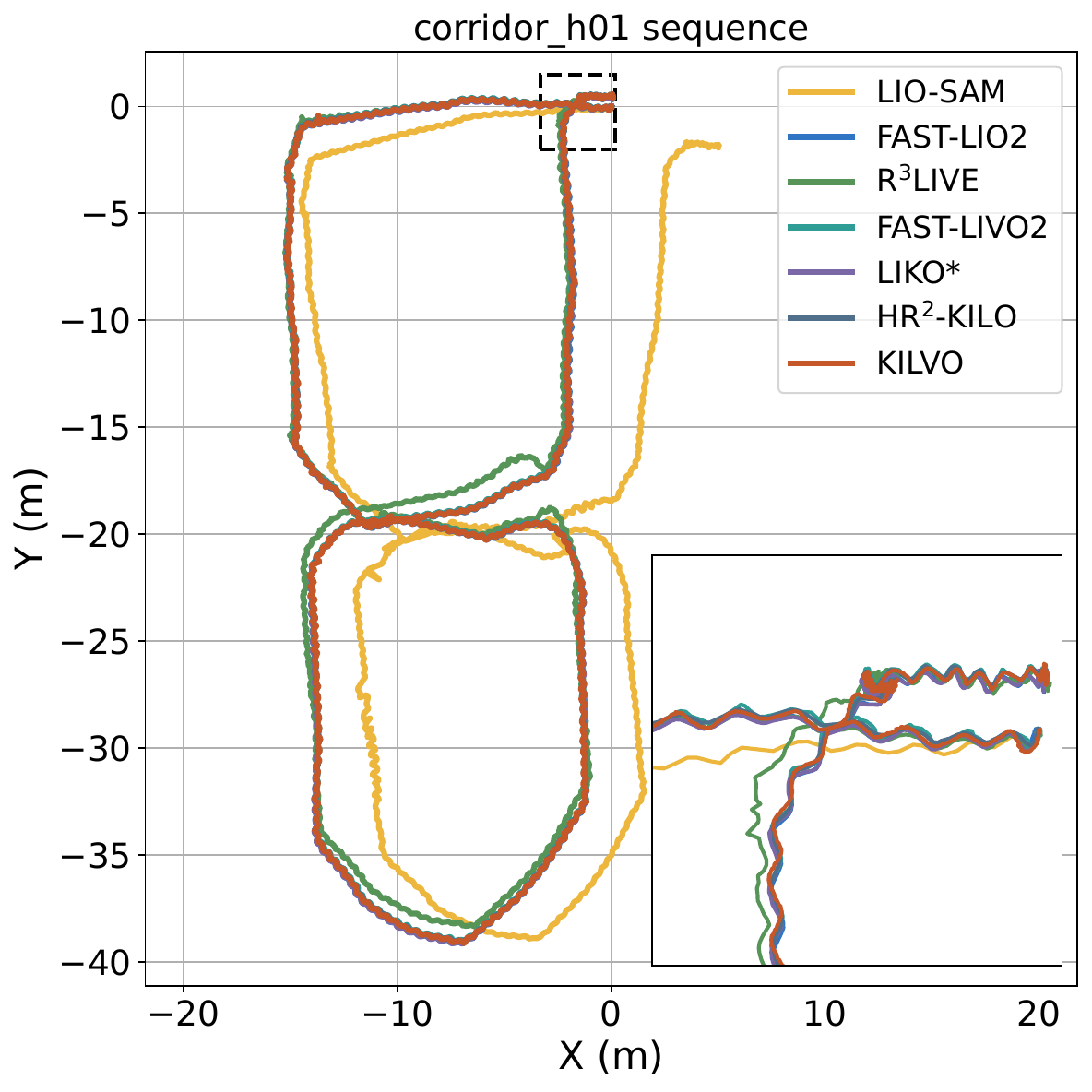}%
      \includegraphics[height=3.815cm]{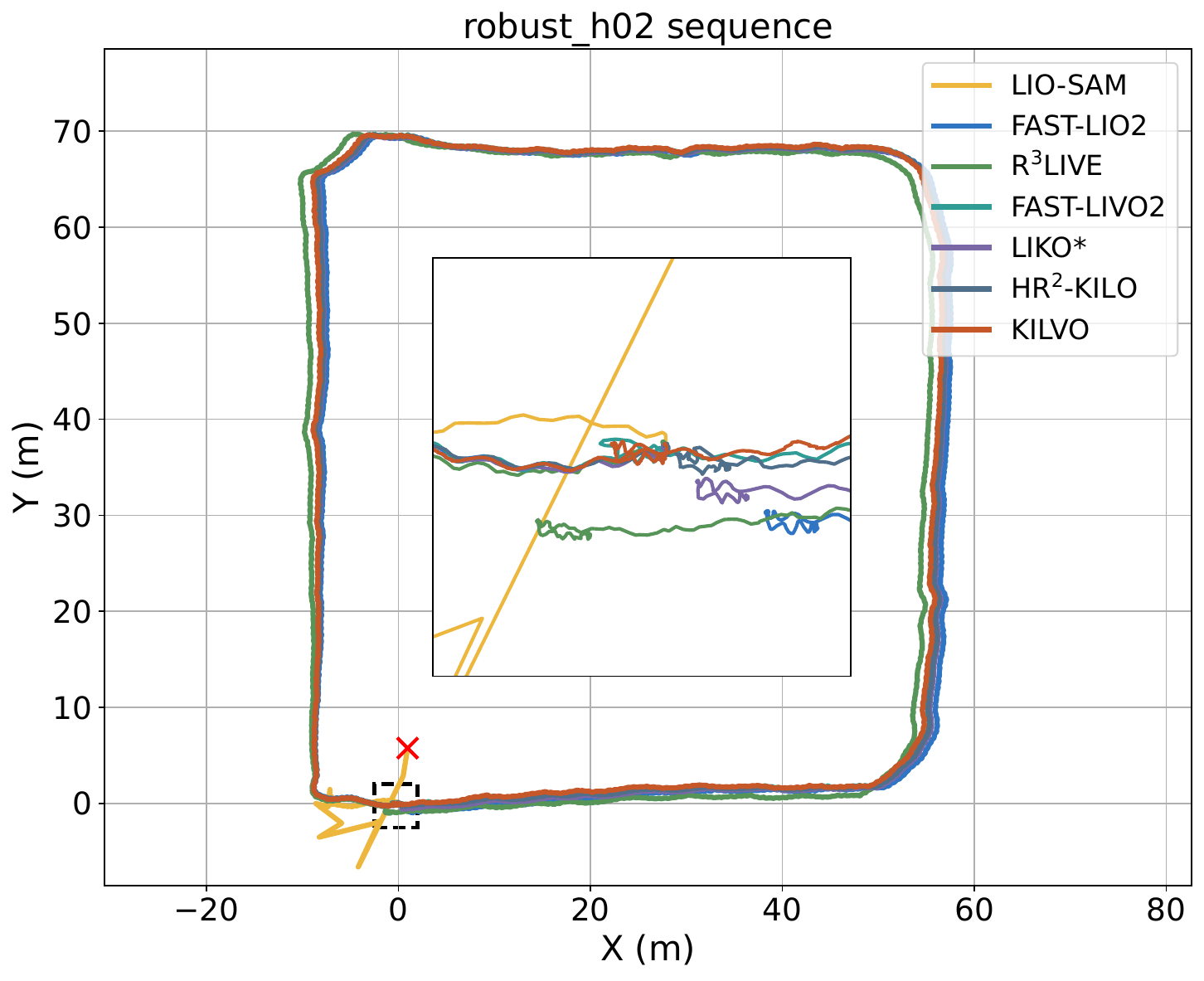}
      \caption{Estimated trajectories of the tested methods in real-world experiments on corridor\_h01 and robust\_h02 sequences.}
      \label{fig_acc_traj}
  \end{figure}

  For more comprehensive evaluation in real-world experiments, we additionally include R$^3$LIVE \cite{R3LIVE}, LIVO configuration of FAST-LIVO2, and KILVO variants.
  The former two extend the benchmark to cover diverse SOTA systems, and the latter enables ablation studies of KILVO under different modalities.
  For all the real-world sequences, the robot is finally returned close to the starting point, where the designed mechanical module is used to measure the ground-truth relative translation.
  
  The real-world experimental results on accuracy are reported in Table \ref{tab:Xaccuracy}.
  It can be seen that our method delivers the best overall accuracy with the average end-to-end translation error of 0.0145 m. 
  KILVO achieves the highest accuracy in {8} of the 15 sequences, and ranks second on 2 of the remaining sequences.
  LIO-SAM fails on partial sequences because ground impacts disrupt the feature association.
  The other methods exhibit varying levels of accuracy depending on the environments and gaits.
  Among these, FAST-LIVO2, along with KILVO and its variants, demonstrates more stable performance.
  R$^3$LIVE tends to degrade in accuracy during aggressive motion, such as on ``run" sequences.
  LIKO and HR$^2$-KILO suffer accuracy loss when leg kinematic constraints are unavailable or incorrect. 
  The former cannot enforce kinematic constraints without contact sensors, while the latter's contact estimation module suffers degradation on challenging terrains.
  This reduction is more pronounced in ``robust'' sequences with frequent ground impacts, unstructured scenes, and difficult terrains.
  Some trajectory comparisons can be seen in Fig. \ref{fig_acc_traj}.
  Benefiting from the stable contact estimation module and the asynchronous-sequential hybrid ESIKF, KILVO tightly couples leg kinematic, inertial, LiDAR, and visual information, achieving superior stability and accuracy. 

  \begin{figure}[!t]
    \centering
    	{\includegraphics[width=0.48\linewidth]{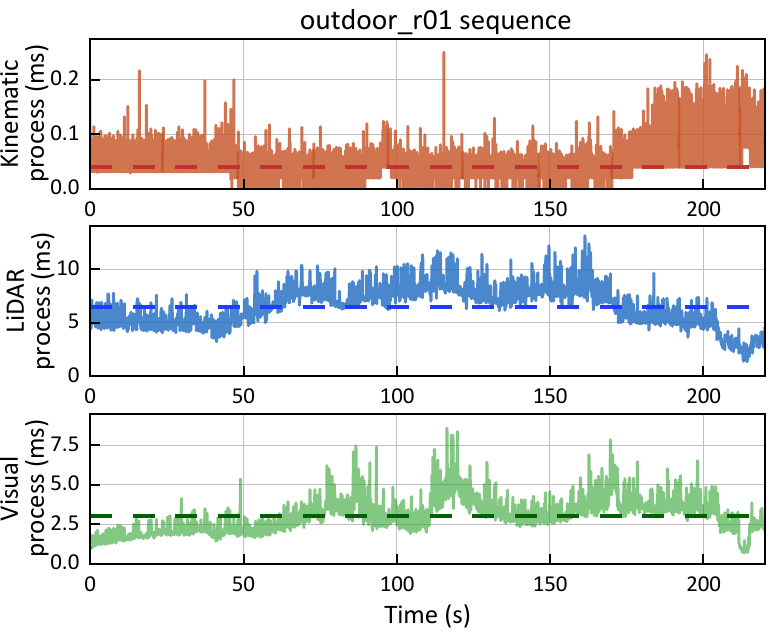}}
			{\includegraphics[width=0.48\linewidth]{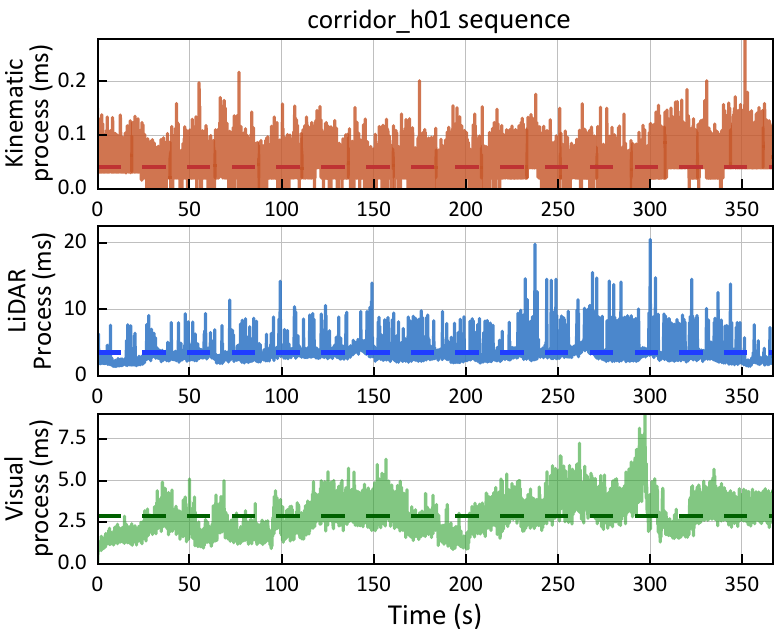}}
    \caption{Time usage per frame for KILVO processing different measurements in state estimation.}
    \label{fig_efficiency}
  \end{figure}

  In ablation studies of KILVO under different modalities, we found that even with a global shutter, image quality often degrades due to the ground impacts. 
  The max exposure time needs to be limited to avoid motion blur, which in turn increases the sensitivity to illumination.
  In addition, a stable contact state ensures the effectiveness of kinematic constraints, thereby contributing positively to the system.
  However, it is affected by the unavoidable factors such as slippages and movable ground tiles.
  Although not all the sensors can contribute positively under any conditions, our system tightly couples them to resist such compound disturbances, demonstrating its robustness.
  For example, in the sequences of ``robust\_h01'' and ``robust\_h02'', KILVO still returns to the starting point with the end-to-end errors around 0.01 m.

\subsection{Efficiency Evaluation}
  In this section, the same configurations are implemented to evaluate the system efficiency.
  % The comparison algorithms exclude LIO-SAM and LIKO*, %as the former fails on partial sequences, and the latter is similar to FAST-LIO2 when not equipped with contact sensors.
  The average time consumption for the complete process is calculated in Table \ref{tab:efficiency1}, which for KILVO comprises several asynchronous phases and a sequential phase.
  R$^3$LIVE has the largest time consumption (about 137 ms) as its workload for maintaining the map through Bayesian update increases with higher resolutions.
  FAST-LIO2 and FAST-LIVO2 exhibit superior real-time performance with the efficiency about 10 ms, but the output rate is limited by the LiDAR/camera measurement frequency.
  HR$^2$-KILO achieves a higher output rate with about 27 ms average time consumption. 
  Except for LiDAR processing, additional computational load is mainly from the contact estimation, which requires high-frequency retrieval of map points. 
  Although the proposed system does not achieve the best efficiency, it fuses four different sensors and maintains a competitive average processing time of 14 ms with 1 kHz output.

  We further report the time consumption of KILVO for processing different measurements in state estimation, as shown in Fig. \ref{fig_efficiency}.
  The time cost for each kinematic processing is generally below 0.2 ms, fulfilling real-time operation at 1 kHz.
  For the sequential phase, thanks to raw points direct registration and sparse image alignment, LiDAR processing typically takes less than 10 ms, and visual processing is lower at around 5 ms.

  \begin{figure*}[!t]
    \centering
    \includegraphics[width=\linewidth]{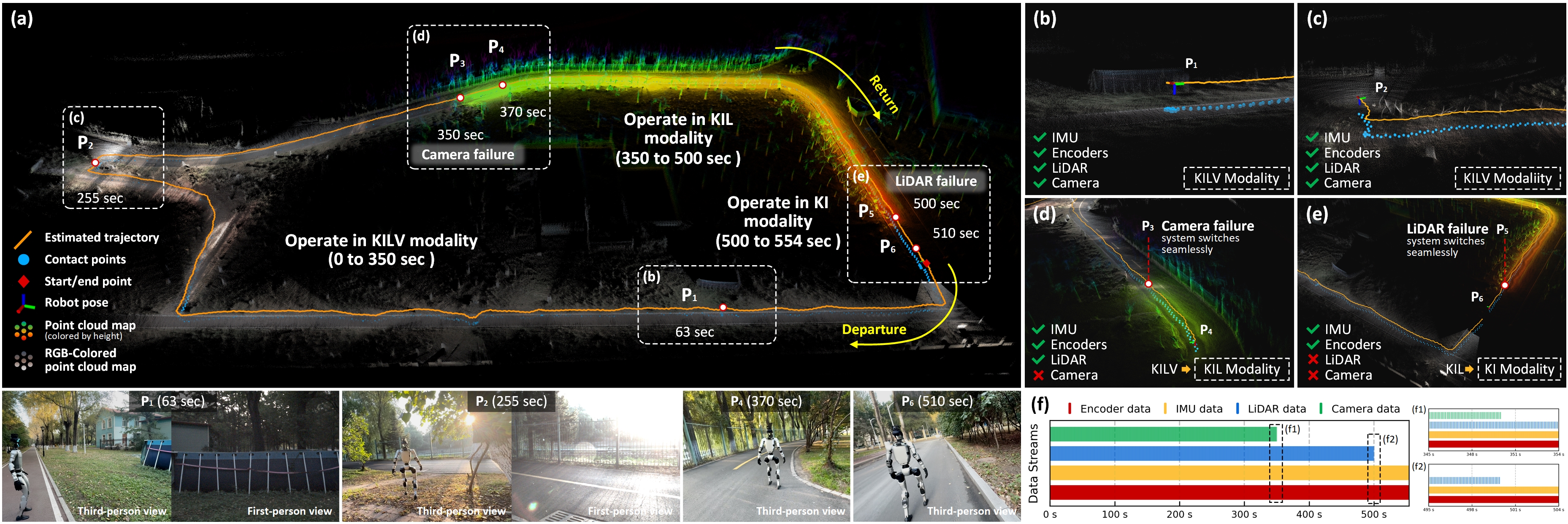}
    \caption{Experimental results under unrecoverable {sensor failure} cases. (a) Estimated trajectory and mapping results of KILVO in robust\_h01* (data corruption version), with snapshots at positions p\textsubscript{1} $\sim$ p\textsubscript{6} shown in either third-person or first-person view. (b) and (c) Experimental details under KILV modality. (d) Modal {adaptation} when the camera fails. (e) Modal {adaptation} when the LiDAR fails. (f) Sensor data streams in this experiment.}
    \label{fig_robust_1}
  \end{figure*}      %--------------------------------------------------------- BIG ROBUST FIG END
  \begin{table}[t!]  %--------------------------------------------------------- TABLE BEG
    \caption{Average Time Consumption (Milliseconds)} \label{tab:efficiency1}
    \centering
    \begin{threeparttable}
      \begin{tabularx}{\linewidth}{p{1.8cm} *{5}{>{\centering\arraybackslash}X}}
        \toprule
        {Sequence}        & {FAST-LIO2} & {R$^3$LIVE} & {FAST-LIVO2} & {HR$^2$-KILO} & {KILVO}\\
        \midrule
        \multicolumn{2}{l}{LIKO Dataset\tnote{a}} & \\
        \midrule
        {fwd\_bwd}        & 11.99 & \textemdash\tnote{b} & 7.64 & 21.40 &  9.71\\
        {square\_walk}    & 17.92 & \textemdash & 8.26 & 22.81 &  10.77\\
        {square\_walk\_l} & 13.84 & \textemdash & 8.13 & 21.87 &  9.93\\
        {walk\_in\_place} & 14.90 & \textemdash & 8.53 & 23.95 &  10.48\\
        {up\_slope}       & 8.99  & \textemdash & 4.81 & 12.48 &  9.30\\
        \midrule
        \multicolumn{2}{l}{KILVO Dataset} & \\
        \midrule
        {corridor\_r01}  & 7.43  & 163.09  & 9.01      & 18.21 & 15.04\\
        {spin\_r01}      & 7.78  & 188.12  & 9.35      & 29.98 & 15.25\\
        {outdoor\_r01}   & 10.27 & 142.06  & 11.18     & 29.47 & 15.55\\
        {outdoor\_r02}   & 10.08 & 121.46  & 11.68     & 31.24 & 19.05\\
        {outdoor\_r03}   & 10.97 & 144.12  & 9.61      & 30.75 & 15.94\\
        {corridor\_h01}  & 7.26  & 130.96  & 9.03      & 21.80 & 12.33\\
        {spin\_h01}      & 6.61  & 147.47  & 8.03      & 34.53 & 12.62\\
        {outdoor\_h01}   & 10.51 & 119.84  & 12.28     & 29.70 & 15.81\\
        {outdoor\_h02}   & 10.70 & 117.58  & 12.24     & 31.95 & 15.61\\
        {outdoor\_h03}   & 9.98  & 134.57  & 11.23     & 36.32 & 15.15\\
        {outdoor\_h04}   & 9.50  & 139.48  & 10.44     & 32.38 & 14.59\\
        {run\_h01}       & 6.31  & 112.44  & 7.27      & 26.98 & 12.68\\
        {run\_h02}       & 6.21  & 120.69  & 7.83      & 26.08 & 12.98\\
        {robust\_h01}    & 14.53 & 126.12  & 18.26     & 32.24 & 18.49\\
        {robust\_h02}    & 14.78 & 151.40  & 15.62     & 32.98 & 16.09\\
        \midrule
        {Average}        & 10.53 & 137.29  & 10.02     &27.36 & 13.99\\
        \bottomrule
      \end{tabularx}
        \begin{tablenotes}[para,flushleft]
          \footnotesize \item[a]  Visual components of FAST-LIVO2 and KILVO are disabled on the dataset.\\
          \footnotesize \item[b] -- denotes that the sequences use a 360° mechanical LiDAR and do not produce the image data, which are not inherently compatible with R$^3$LIVE.
          
        \end{tablenotes}
    \end{threeparttable}
  \end{table} %---------------------------------------------------------------- TABLE END
\subsection{Robustness Performance}
  \subsubsection{Unrecoverable Sensor Failure}
  Here, we validate the robustness against unrecoverable sensor failures during system operation.
  The ``robust\_h01'' sequence is used in this experiment, which lasts over 9 minutes and features unstructured scenes, variable-lighting conditions, and diverse terrains.
  To simulate {data-stream interruption}, the camera and LiDAR data are sequentially filtered out during the task and not recovered.
  
  This process is shown in Fig. \ref{fig_robust_1}.
  When the humanoid robot departs from the starting point, our system operates in the full modality with all the sensors functioning until 350 s.
  In this period, the robot experiences continuous ground impacts while navigating changes in terrain and illumination.
  From 350 s, image is completely lost until the end.
  KILVO seamlessly transitions to the KIL modality at 350 s, maintaining continuous localization and mapping [see Fig. \ref{fig_robust_1}(d)].
  Moreover, the mapping output is from RGB-colored maps to geometric maps due to the loss of the image texture. 
  At 500 s, LiDAR data is also cut off, leading the system to degrade again and seamlessly adapt to the KI modality [Fig. \ref{fig_robust_1}(e)].
  Until returning to the starting point, our system {maintains} contact estimation and localization {using only inertial and kinematic data}.
  KILVO completes {this challenging} task with an end-to-end translation error of 0.3338 m.
  Compared with the 0.0130 m of the full modality, the degradation {mainly comes from proprioceptive drift, including non-ideal contacts, kinematic modeling errors, and approximate ground patches.}
  This trade-off is acceptable given the improved robustness, {but prolonged loss of exteroception still limits the system global accuracy.}

  \subsubsection{Sensor Failure \& Recovery}
  We also perform experiments to evaluate KILVO under different sensor failure and recovery cases.
  The corrupted version of ``robust\_h02'' is used in this section, named ``robust\_h02*'', which lasts a total of 375 s.
  The partial sensor data segments in this sequence are filtered out. 
  Specifically, the {kinematic} data is lost from 60 to 70 s, LiDAR is lost from 150 to 160 s, the image is lost from 230 to 240 s, the remaining data is intact.
  \begin{figure}[!t]
    \centering
    \includegraphics[width=\linewidth]{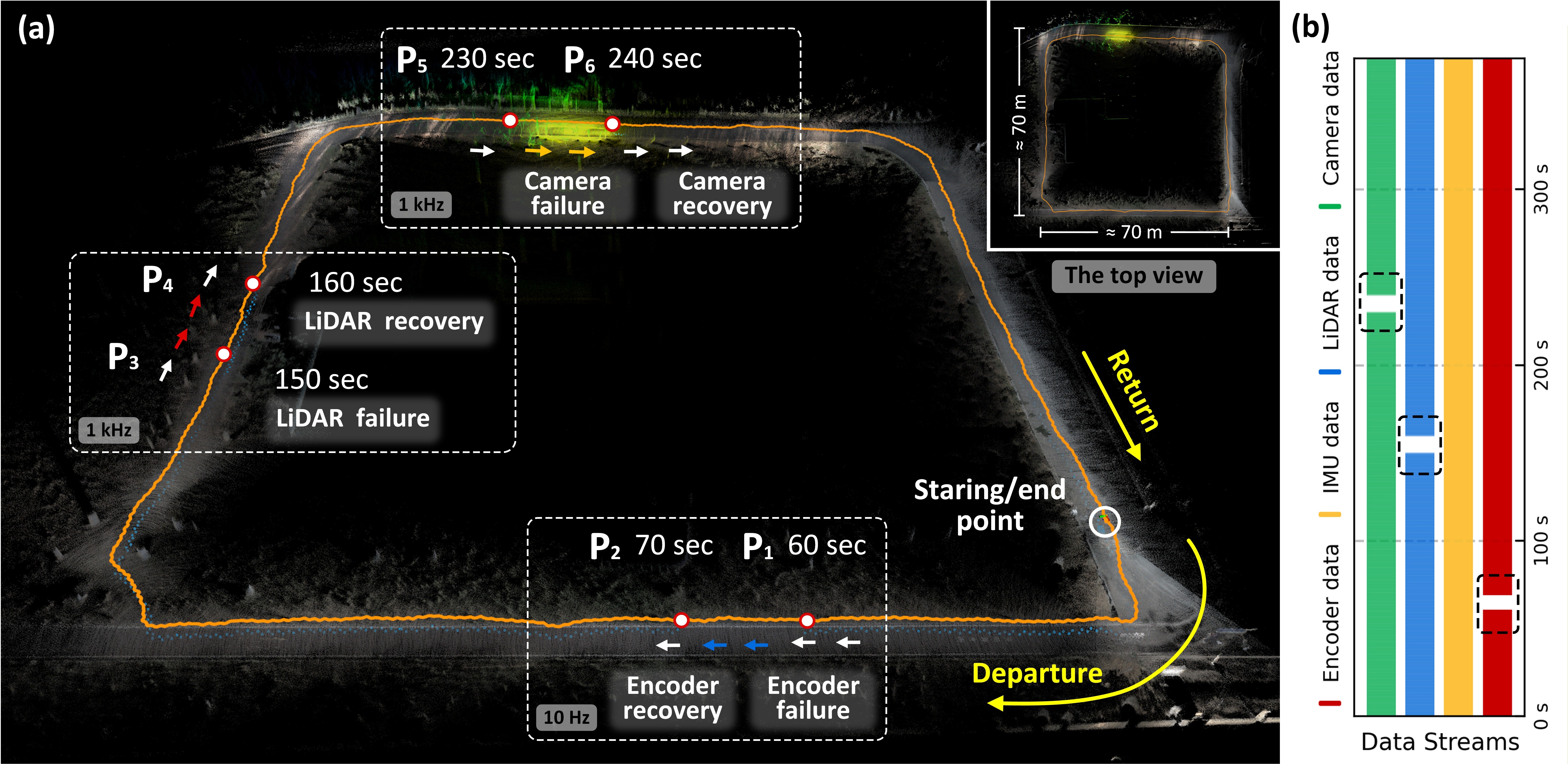}
    \caption{Experimental results under {sensor} failure and recovery cases. (a) Estimated trajectory and mapping results of KILVO on robust\_h02*(data corruption version). (b) Sensor data streams in this experiment.}
    \label{fig_robust_2}
  \end{figure}

  The experimental results are shown in Fig. \ref{fig_robust_2}.
  During the initial period, the system operates in the full modality until 60 s.
  The system detects the encoder failures by checking data streams and adapts to LIV modality.
  The localization and mapping do not cease, but the output rate will be reduced from 1 kHz to 10 Hz until the encoders recover, after which the leg kinematic constraints are enforced again [Fig. \ref{fig_robust_2} (b)].
  From 150 to 160 s, the system estimates states in KI modality without mapping due to LiDAR failure.
  This also allows our system to re-fuse LiDAR when it is recovered, avoiding drastic jumps in localization.
  KILVO is degraded to KIL modality at 230 s, during which the system performs asynchronous state estimation, and the RGB point cloud is replaced by the raw point map.
  Then, the system reverts to the full modality and completes this task with an end-to-end error of 0.0165 m.
  Benefiting from the tightly coupled multisensor framework with the multiple modal adaptation, KILVO exhibits sufficient robustness to tolerate the sensor failures.

\begin{figure}[!t]
    \centering
      {\includegraphics[width=\linewidth]{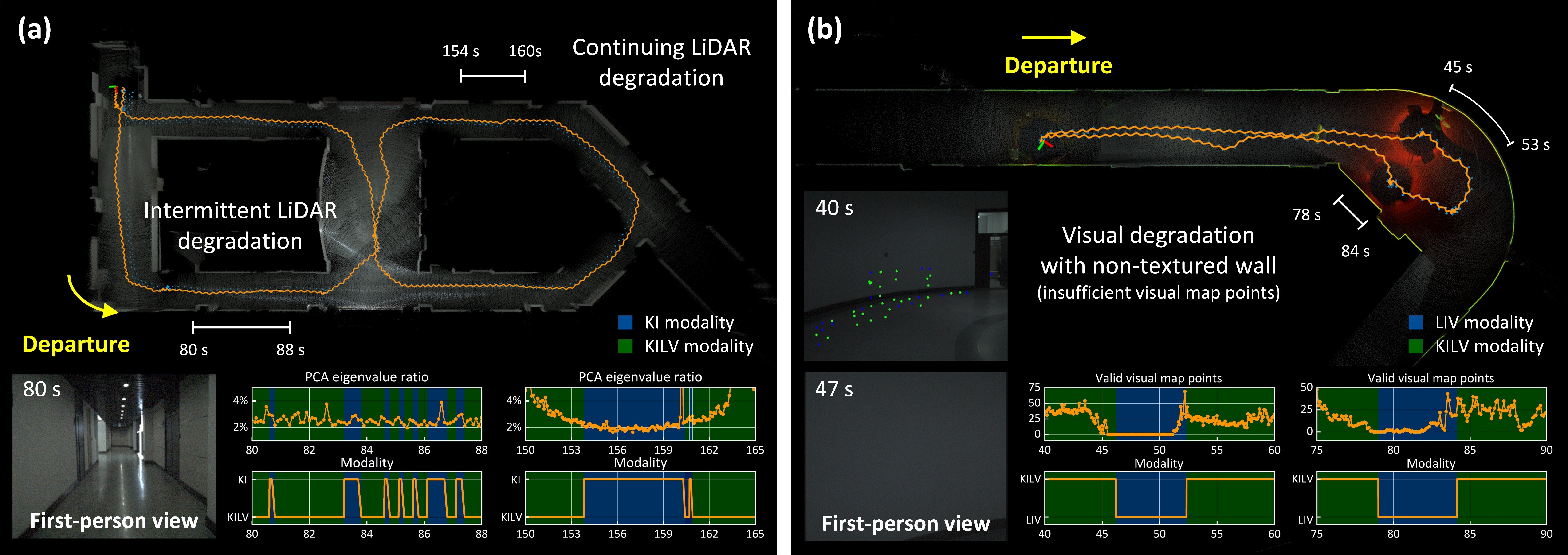}}
    \caption{{Experimental results under measurement degradation. (a) LiDAR degradation in a corridor. (b) Visual degradation near a non-textured wall.}}
    \label{fig_deg}
  \end{figure}

  {
  \subsubsection{Measurement Degradation}
  To further test multimodal adaptation under degraded measurements, LiDAR and visual degradation are considered.
  For each LiDAR frame, the degradation is detected by the PCA-based eigenvalue ratio.
  As shown in Fig. \ref{fig_deg}(a), KILVO intermittently transitions modalities to handle the degraded point clouds (after 80 s) and continuously relies on the KI modality during 154--160 s.
  Fig. \ref{fig_deg}(b) shows that valid visual map points nearly vanish as the robot approaches a non-textured wall, the system transitions to KIL without visual updates.
  KILVO remains robust to low-quality measurements and flexibly supports different degradation indicators to drive the modality adaptation.
  }
\section{Conclusion}
This article presented KILVO, a multisensor fusion framework for localization and mapping on humanoid robots.
The measurements from leg kinematics, IMU, LiDAR, and camera are tightly coupled within an asynchronous-sequential hybrid ESIKF, allowing accurate and efficient state estimation with high-rate output.
And the system is elaborately designed to enable seamless multimodal adaptation.
A stable and efficient contact estimation module is also developed and integrated into the framework.
Extensive experiments are conducted on public datasets and in the real world,
 demonstrating that our approach achieves a high level of accuracy and efficiency while delivering strong robustness against sensor degradation and failures.
{In the future, learning-based modules are expected to be flexibly integrated into KILVO for improved adaptability.}
% Compared to the SOTA methods, KILVO is more suitable for humanoid robots as it considers the challenges from both the platform's features and the real-world complexity. 
 

\bibliographystyle{IEEEtran}
\bibliography{IEEEabrv, ./bib/paper}

% \newpage

\end{document}